\documentclass[11pt]{article}
\PassOptionsToPackage{table}{xcolor}

\usepackage[final]{acl}

\usepackage{fontspec}

\usepackage{polyglossia}

\setdefaultlanguage{english}
\setotherlanguage{arabic}

\newfontfamily\arabicfont[
  Script=Arabic,
  Language=Arabic,
  Scale=1.1,
  Path=fonts/,
  Extension=.ttf,
  UprightFont=Amiri-Regular,
  BoldFont=Amiri-Bold,
  ItalicFont=Amiri-Slanted,
  BoldItalicFont=Amiri-BoldSlanted
]{Amiri}

\newfontfamily\arabicfontfamily[
  Script=Arabic,
  Scale=1.1,
  Path = fonts/,
  Extension = .ttf,
  UprightFont = Amiri-Regular,
  BoldFont = Amiri-Bold,
  ItalicFont = Amiri-Slanted,
  BoldItalicFont = Amiri-BoldSlanted
]{Amiri}

\usepackage{latexsym}

\usepackage{amsmath}
\usepackage{amssymb}
\usepackage{amsfonts}
\usepackage{nicefrac}

\usepackage[table]{xcolor}
\usepackage{array}
\usepackage{booktabs}
\usepackage{multirow}
\usepackage{makecell}
\usepackage{adjustbox}
\usepackage{longtable}
\usepackage{ragged2e}

\usepackage{graphicx}
\usepackage{subcaption}
\usepackage{float}
\usepackage{placeins}
\usepackage{afterpage}

\usepackage{algorithm}
\usepackage{algpseudocode}

\usepackage{mdframed}
\usepackage{tcolorbox}
\tcbuselibrary{skins}
\usepackage{soul}
\usepackage{pifont}
\usepackage{parskip}

\usepackage{url}
\usepackage{hyperref}

\definecolor{headernavy}{HTML}{1a2e4a}
\definecolor{boxbg}{HTML}{eef2f7}

\definecolor{sblue}{HTML}{87CEEB}
\definecolor{teal}{HTML}{2A9D8F}
\definecolor{navy}{HTML}{1565C0}
\definecolor{mango}{HTML}{F4C430}
\definecolor{tang}{HTML}{F07C00}
\definecolor{crim}{HTML}{C0392B}

\colorlet{sblueL}{sblue!32}
\colorlet{crimL}{crim!32}
\colorlet{tealL}{teal!32}
\colorlet{mangoL}{mango!32}

\definecolor{noteblue}{RGB}{30,100,180}
\definecolor{notegray}{RGB}{100,100,100}
\definecolor{highlight}{RGB}{255,245,200}

\newmdenv[
  backgroundcolor=highlight,
  linecolor=orange!60,
  linewidth=1pt,
  innerleftmargin=10pt,
  innerrightmargin=10pt,
  innertopmargin=8pt,
  innerbottommargin=8pt,
]{keybox}

\title{Bridging the English-Arabic Medical Knowledge Gap: Targeted Low-Rank Adaptation via Causal Layer Selection}

\author{
Chaimae Abouzahir$^1$\quad\quad
Musa Khan$^1$\quad\quad
Hala Ali-Hassan$^1$\quad\quad
Congbo Ma$^1$\\
\bfseries  Khaled Saleh$^2$\quad\quad
Yousra Sadqi$^2$\quad\quad
Jihad Mallat$^2$\quad\quad
Walid Al-Eisawi$^1$\\
\bfseries Nizar Habash$^1$\quad\quad
Farah E. Shamout$^1$
\\[0.5em]
$^1$New York University Abu Dhabi \\
$^2$Cleveland Clinic Abu Dhabi \\
\texttt{ca2627@nyu.edu}
}

\begin{document}
\maketitle
\begin{abstract}
Large Language Models (LLMs) perform strongly in English medical tasks but degrade substantially in Arabic, a gap widely attributed to limited training data. We systematically investigate this assumption via tuned lens probing and causal activation patching, and find that Arabic medical knowledge is present in intermediate model representations but fails to surface at the output. This mechanistic insight motivates a targeted adaptation strategy. Rather than fine-tuning the full network, we propose Targeted Low-Rank Adaptation (TLoRA), restricted to the layer window where cross-lingual representations diverge, upstream of the output layers where the failure manifests. We evaluate TLoRA on multiple-choice medical QA, where our approach outperforms full-network LoRA, zero-shot, and few-shot baselines. We further evaluate it on short-answer generation and multi-turn clinical dialogue, where it performs competitively without the need for task-specific finetuning. We additionally introduce AraClinicDialog, a clinician-constructed Arabic medical dialogue benchmark in MSA with validated variants across four Arabic dialects. Together, these contributions demonstrate that mechanistic diagnosis can serve as a practical guide for targeted adaptation in underrepresented-language medical LLMs. \footnote{Code available at \url{https://github.com/nyuad-cai/TLoRA-Adaptation}.}


\end{abstract}

\section{Introduction}

The rise of Large Language Models (LLMs) has led to rapid progress in healthcare, with systems demonstrating strong performance across medical question answering, clinical language understanding, and real-world diagnostic prediction \cite{pmlr-v219-wang23c, nazi2024, jiang2023health}. However, this progress has been defined and measured mainly in English \citep{betteraskinenglish, joshi-etal-2020-state}. Pretraining corpora remain heavily English-dominated, even for nominally multilingual models \citep{touvron2023llamaopenefficientfoundation, workshop2023bloom176bparameteropenaccessmultilingual}, and the benchmarks 
used to evaluate medical capability are similarly English-centric \citep{qiu2024multilingual, pmlr-v174-pal22a, jin-etal-2019-pubmedqa}. Outside English, prior work shows that LLM performance degrades on medical tasks \cite{wang2024apollolightweightmultilingualmedical, betteraskinenglish}. Importantly, the mechanisms underlying this degradation remain poorly understood and insufficiently addressed, particularly in low-resource and underrepresented languages.

Arabic, spoken by over 400 million people, remains underrepresented in medical LLM training and evaluation due to the scarcity of high-quality domain-specific data \citep{ statista_languages_2023, daoud2026medarabench}. Beyond data availability, Arabic is morphologically rich, and its diglossic structure spans Modern Standard Arabic (MSA) and dialectal variants that lack written standards and resources \citep{habash2010introduction, Moaiad2024ChallengesIN}. These factors make Arabic a particularly challenging setting for cross-lingual generalization in medical LLMs. Despite growing efforts in Arabic medical NLP, performance gaps persist and their causes remain underspecified \citep{abouzahir-etal-2026-cross}.

Existing approaches to improving Arabic medical performance largely treat this as a data problem. Domain-specific efforts, such as BiMediX \citep{pieri-etal-2024-bimedix}, rely on translating English medical data and fine-tuning on the result. However, this approach inherits English-centric clinical norms and biases while lacking native Arabic grounding, and recent work shows that fine-tuning on Arabic medical data can even degrade performance on several benchmarks \citep{saadi2025bridging}. At the same time, general-purpose Arabic LLMs, including Jais and ALLaM, are not designed for medical reasoning and perform poorly on medical domain benchmarks \citep{pmlr-v298-daoud25a}. Broader multilingual adaptation methods, such as representation alignment and encoder-bridging approaches \citep{zhao2025lens, yoon-etal-2024-langbridge, huang2024mindmerger}, offer only limited gains and assume the bottleneck is data quantity or language coverage, leaving the model's internal failure mechanism unexamined.


Crucially, all of these approaches treat the model as a black box: none ask where in the model the failure occurs or why Arabic queries fail to elicit  knowledge the model demonstrably possesses. We identify that Arabic medical failure in LLMs is a knowledge-routing breakdown 
rather than a knowledge deficit. More specifically, the model answers correctly in English on approximately 30\% of questions where it fails on an identical Arabic query (Mistral-Small-3.2-24B on MedAraBench; see \S\ref{sec:mechanistic}). Using tuned lens probing, causal activation patching, and KL divergence profiling, we localize this breakdown to a specific layer window and derive a targeted 
adaptation strategy, whose design follows directly from the mechanistic evidence. Our contributions are as follows:

\begin{itemize}

 \item To the best of our knowledge, we provide the first mechanistic analysis of Arabic medical failure in LLMs, identifying a \textbf{knowledge-routing failure} specific to a layer window that motivates our adaptation design.

 \item We propose \textbf{layer-targeted LoRA}, denoted as TLoRA, with
a cross-lingual alignment objective whose window and probe layer are both derived from mechanistic evidence.

 \item We present \textbf{AraClinicDialog}, a new clinician-constructed Arabic medical 
dialogue benchmark in MSA with validated variants across four Arabic dialects.

\end{itemize}

\section{Related Work} 
\subsection{LLMs in Healthcare}

Clinical question answering benchmarks have become the primary measure of progress for medical LLMs, with proprietary systems such as GPT-4 and Med-PaLM 2 now achieving near-expert performance and medical-domain open models such as Meditron increasingly approaching the same ceiling \citep{singhal2023llmclinical, nori2023capabilitiesgpt4medicalchallenge,  singhal2025medpalm2, chen2023meditron70bscalingmedicalpretraining}. These benchmarks, however, are designed around English-language clinical data and evaluation conventions, establishing a performance standard whose underlying assumptions are English-first. 
Efforts to extend evaluation beyond English exist but remain constrained, relying on translation from other source languages \citep{Alonso_2024}.

While MCQA benchmarks offer a scalable and reproducible measure of medical knowledge, they remain an insufficient basis for assessing clinical competence, as the format is susceptible to artifacts and structural cues \citep{cocchieri-etal-2026-remedqa}.
Recent work has therefore extended evaluation to open-ended generation and clinical dialogue. For example, Med-PaLM 2 introduced physician preference judgments over long-form answers, AMIE evaluated diagnostic reasoning through blinded OSCE-style consultations, and HealthBench and MedHELM formalized rubric-based scoring over free-form outputs \citep{singhal2025medpalm2, tu2025amie, arora2025healthbenchevaluatinglargelanguage, bedi2026medhelm}. Like their MCQA counterparts, however, these frameworks are developed and validated primarily in English.

 Cross-lingual degradation in medical LLMs has been documented consistently across languages and evaluation paradigms. In MCQA settings, MedExpQA reports roughly a ten-point accuracy drop across European languages \citep{Alonso_2024}, while multilingual medical models such as Apollo and Apollo-MoE still show a fifteen-point gap on Arabic after multilingual adaptation. \citep{wang2024apollolightweightmultilingualmedical, zheng2025efficiently}. Beyond MCQA, Jin et al. \cite{betteraskinenglish} document correctness, consistency, and verifiability disparities in open-ended healthcare queries.  In each case, the degradation is attributed to data scarcity or insufficient multilingual pretraining, leaving open whether the failure instead reflects an inability to retrieve knowledge already present in the model. 

\subsection{Arabic Medical LLMs} 

At the intersection of Arabic language and clinical medicine, LLM development has been largely absent. Models trained specifically on Arabic such as Jais and Allam demonstrate strong performance across standard NLP tasks but underperform on clinical benchmarks \citep{pmlr-v298-daoud25a}. A first wave of dedicated work has begun to address this with new benchmarks establishing evaluation grounded in native Arabic medical material \citep{pmlr-v298-daoud25a, daoud2026medarabench}, and BiMediX representing the first purpose-built Arabic medical LLM through bilingual fine-tuning on translated clinical data \citep{ pieri-etal-2024-bimedix}.

Model adaptation efforts share some limitations. First, training and evaluation data are derived primarily from English translation pipelines rather than native Arabic clinical sources. All existing work frames the Arabic medical capability gap as a problem of data coverage and bilingual supervision rather than one of representational access. One study \cite{abouzahir-etal-2026-cross} confirms through cross-lingual empirical analysis that the gap is consistent and domain-specific, yet stops short of a mechanistic account.

\subsection{Cross-Lingual Transfer and Representation Alignment}

The mechanistic basis for this gap has been established in the broader multilingual literature. Multilingual LLMs process non-English inputs through an English-centered latent pathway, with cross-lingual prediction failures concentrating precisely at the middle layers where alignment to English breaks down \citep{wendler-etal-2024-llamas, schut2025do}. This failure is causal, patching English hidden states at those layers recovers correct predictions in the majority of failure cases, and disproportionately affects knowledge retrieval rather than abstract reasoning, implicating representational access to stored knowledge as the primary bottleneck \citep{ravisankar-etal-2026-map, hu-etal-2025-large-language, ifergan-etal-2025-beneath}. This misalignment is measurable through alignment scores between parallel representations \citep{kargaran-etal-2025-mexa, hammerl-etal-2024-understanding}, and is especially severe in typologically distant, morphologically complex languages, making Arabic a critical and underexplored test case.

The dynamics of cross-lingual knowledge transfer in domain adaptation have only recently been studied. For example, Kobayashi et al. \cite{kobayashi-etal-2025-leveraging} show that English biomedical corpora support low-resource Japanese medical adaptation but that transfer depends sensitively on corpus composition, while Zhao et al. \cite{zhao-etal-2026-tracing} trace how domain facts are memorized and generalized during multilingual medical adaptation, finding that transfer remains challenging even under high-quality bilingual training. Both studies, however, analyze transfer at the behavioral level, through performance curves and learning dynamics, and examine a relatively high-resource language pair. Our work extends this line to Arabic medicine, a typologically distant and morphologically complex setting, and connects behavioral failure to the representational mechanism: whether Arabic hidden states remain aligned enough to access English-anchored medical knowledge.

\begin{figure*}[!t]
    \centering
    \includegraphics[width=\textwidth]{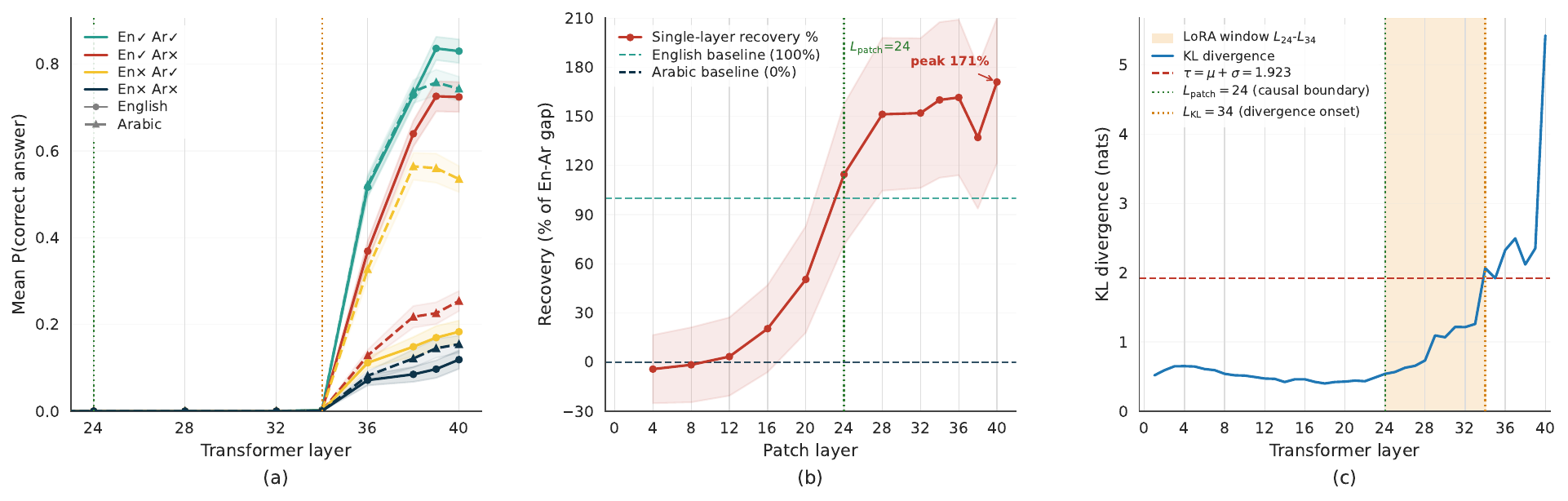}
    \caption{
        Mechanistic motivation for targeted adaptation of Mistral-Small-3.2-24B:
        \textbf{(a)} tuned lens probing,
        \textbf{(b)} causal activation patching, and
        \textbf{(c)} cross-lingual KL divergence profile.
    }
    \label{fig:mech_motivation}
\end{figure*}

Existing adaptation methods fall into three families, none of which addresses the representational root cause identified above. Relearning methods \citep{cui2024efficienteffectivetextencoding, wang2023huatuotuningllamamodel} improve target-language performance but erode cross-lingual structure, inducing catastrophic forgetting in zero-shot generation \citep{vu-etal-2022-overcoming} and proximity-dependent forgetting around injected medical concepts \citep{liu-niehues-2025-conditions, zhou2026investigating}. External bridge methods route non-English reasoning through cross-lingual prompts, trainable bridging parameters, or auxiliary multilingual encoders \citep{huang2023not, yoon-etal-2024-langbridge, huang2024mindmerger}, avoiding full relearning but introducing external components that require additional training and do not leverage the model's existing English domain representations.

To the best of our knowledge, no prior method combines a mechanistic diagnosis of domain-specific knowledge access failure with a targeted adaptation strategy that uses the model's own English hidden states as an alignment anchor, in Arabic or in any other language.

\section{Mechanistic Motivation}
\label{sec:mechanistic}

We initially evaluate Mistral-Small-3.2-24B on the MedAraBench test set~\citep{daoud2026medarabench}, where it achieves 57.8\% in English and 50.8\% in Arabic, using the prompt in Appendix Figure ~\ref{fig:prompt-task1-zero}.  The English set is a machine translation of the original Arabic questions via the Google Translate API, constructed specifically for this diagnostic comparison. Among English-correct questions, 29.6\% fail on the identical Arabic query. This implies that the model demonstrably possesses the knowledge but fails to surface it. We ask if this reflects absent knowledge or a failure of access.

To answer this, we use three complementary mechanistic probes. Tuned-lens probing~\citep{belrose2023tunedlens} and causal
activation patching~\citep{meng2022locating} test whether the correct answer is encoded mid-network and whether replacing Arabic
representations with English ones recovers it, respectively. A cross-lingual KL divergence profile~\citep{wendler-etal-2024-llamas} then
locates where English and Arabic diverge, setting the adaptation window used in \S\ref{sec:window}.

\paragraph{Tuned Lens. } We applied tuned lens probing to track the correct answer probability layer-by-layer across parallel English and Arabic forward passes. Figure~\ref{fig:mech_motivation} (a) shows the correct answer accumulates through intermediate Arabic layers, reaching English levels by mid-network, then collapses before the output; the both-wrong quadrant remains flat throughout. This rules out a general decoding deficit such that the knowledge is encoded but not committed.

\paragraph{Causal Activation Patching. } We apply causal activation patching to verify this causally, replacing Arabic hidden states with English counterparts one layer at a time and measuring gap recovery. Figure~\ref{fig:mech_motivation} (b) shows that recovery exceeds 80\% at a single injection layer $L_{\mathrm{patch}} = 24$, with peak recovery reaching 171\%. Layers below this threshold yield near-zero recovery. This suggests that the failure is both causal and precisely localized. 

\paragraph{KL Divergence Profile. }
We compute the cross-lingual KL divergence profile to identify the upstream source of the breakdown. Figure~\ref{fig:mech_motivation}(c) points to divergence rising sharply at $L_{\mathrm{KL}} = 34$, after which English and Arabic trajectories become irreconcilable.
Patching and KL profiling together identify two mechanistically salient boundary layers, $L_{\text{patch}}=24$ and $L_{\text{KL}}=34$, marking, respectively, the onset of causal recoverability and the onset of sharp cross-lingual divergence, and defining the boundary signals for the candidate adaptation windows in \S\ref{sec:window}.



\section{Methodology}

Our method has two mechanistically-grounded design choices. The first identifies which layers to apply LoRA to, and the second determines at which layer to apply the cross-lingual alignment. Both are determined from the model's own internal signals, the causal boundary $L_\text{patch}$ and the divergence onset $L_\text{KL}$ identified in \S\ref{sec:mechanistic}, before any training begins and fixed 
thereafter. 

\subsection{Mechanistically-Guided Window Selection}
\label{sec:window}

The two boundary layers identified in \S\ref{sec:mechanistic} partition the network into three contiguous regions, generating a small, exhaustive set of candidate adaptation windows shown in Table~\ref{windows}.

\begin{table}[ht]
\centering
\small
\begin{tabular}{lll}
\toprule
Window & Layers & Region \\
\midrule
$W_1$ & L1--$L_\text{patch}$             & below causal boundary \\
$W_2$ & $L_\text{patch}$--$L_\text{max}$ & causal window \\
$W_3$ & L1--$L_\text{KL}$                & below divergence onset \\
$W_4$ & $L_\text{KL}$--$L_\text{max}$    & active zone only \\
$W_5$ & L1--$L_\text{max}$               & full model \\
\bottomrule
\end{tabular}
\caption{Candidate LoRA adaptation windows.}
\label{windows}
\end{table}

Rather than searching arbitrarily over layer subsets, this partition ensures candidates correspond to meaningful network regions defined by causal evidence. We run an independent learning-rate sweep for each window under the full training objective (Appendix \ref{app:lora_lr_search}) and select the window with the best held-out performance. Because $\beta^*$ is derived from initialisation losses before training (see \S\ref{sec:objective}), and LoRA weights are zero at initialisation regardless of window, $\beta^*$ is identical across all five sweeps, ensuring a fair comparison. The causal boundary $L_\text{patch}$ is defined as:
\begin{equation}
  L_\text{patch}
  \;=\; \min\bigl\{\ell : \mathrm{recovery}(\ell) \geq \tau_\text{patch}\bigr\}
\end{equation}
with $\tau_\text{patch} = 0.5$.

\subsection{KL Probe Layer}
\label{sec:kl_probe}

Independently of window selection, we determine the layer at which to apply cross-lingual alignment pressure from the model's own divergence profile. We run parallel-bilingual data forward passes, and at each layer $\ell$ project the final-token hidden state through the frozen RMSNorm and unembedding matrix to obtain vocabulary distributions $\hat{p}_\ell^\text{Ar}$ and $\hat{p}_\ell^\text{En}$. 
The per-layer divergence is:
\begin{equation}
  \mathrm{KL}(\ell)
  \;=\; \mathbb{E}_{(x^\text{Ar},\,x^\text{En})}\!\left[
        D_\text{KL}\!\left(\hat{p}_\ell^\text{En}
                           \,\big\|\,
                           \hat{p}_\ell^\text{Ar}\right)\right]
\end{equation}

The profile partitions naturally into three zones: neutral, ramp, and active, defined by the mean $\mu$ and standard deviation $\sigma$ of $\mathrm{KL}(\ell)$ across layers. The probe layer is the first layer to enter the active zone:
\begin{equation}
  L_\text{KL}
  \;=\; \min\bigl\{\ell : \mathrm{KL}(\ell) \geq \mu + \sigma\bigr\}
\end{equation}
%


\subsection{Training Objective}
\label{sec:objective}

We train with a combined cross-entropy and cross-lingual 
alignment loss:
\begin{equation}
  \mathcal{L}
  \;=\; \mathcal{L}_\text{CE}
        \;+\; \beta^*\,\mathcal{L}_\text{align}
\end{equation}

where $\mathcal{L}_\text{align}$ is the KL divergence between 
English and Arabic logit-lens distributions at $L_\text{KL}$:
\begin{equation}
  \mathcal{L}_\text{align}
  \;=\; D_\text{KL}\!\left(
        \hat{p}_{L_\text{KL}}^\text{En}
        \,\big\|\,
        \hat{p}_{L_\text{KL}}^\text{Ar}
        \right)
\end{equation}

We calibrate $\beta$ from the model at initialization. A single forward pass measures the initial losses $\mathcal{L}_\text{CE}^{(0)}$  and $\mathcal{L}_\text{align}^{(0)}$, and we set:
\begin{equation}
  \beta^*
  \;=\; \frac{\mathcal{L}_\text{CE}^{(0)}}{\mathcal{L}_\text{align}^{(0)}}
\end{equation}

The alignment term requires parallel Arabic--English pairs at every training step. Arabic provides the student distribution, while English, computed without gradient, provides the teacher. The CE term trains on Arabic medical MCQs only.

\section{Experimental Setup}
\subsection{Implementation Details}
Unless otherwise specified, all models are evaluated zero-shot, with task instructions provided through prompting (Appendix~\ref{app:exp-setup}). For all trained adaptation methods, the MedAraBench training split (17,860 training examples and 1,987 validation examples; stratified 90/10 split) is the sole training source. For TLoRA and LoRA v2, the alignment objective uses the same Arabic training examples machine-translated into English via Google Translate API to form parallel pairs. The task and alignment losses are optimized jointly at every training step.

Applying the layer-selection criteria (\S\ref{sec:mechanistic}) to Mistral-Small-3.2-24B identifies two boundaries: $L_\text{patch} = 24$ and $L_\text{KL} = 34$ ($\mu = 0.88$, $\sigma = 1.04$, $\tau = 1.92$). These separate the network into five candidate adaptation windows (Table~\ref{tab:window_ablation}), from which the optimal window is selected via held-out performance on the training split.

\subsection{Evaluation Tasks}
We evaluate across three tasks of increasing complexity: multiple-choice question answering (MCQA), short answer generation, and multi-turn clinical dialogue. MCQA is the primary task our method is optimized for, while the remaining two serve as generalization probes, testing whether adaptation gains transfer without catastrophic forgetting. 

\subsubsection{Multiple-Choice Question Answering}

We evaluate on four native Arabic MCQA benchmarks: MedArabiQ \citep{pmlr-v298-daoud25a}, MedAraBench \citep{daoud2026medarabench}, and the biology and medicine subsets of ArabicMMLU \citep{koto-etal-2024-arabicmmlu} and AraSTEM \citep{boussaha-etal-2025-3lm}. All are derived from regional medical examinations in MSA. We retain only questions with at least four answer 
options for consistency. Models predict the correct letter and are evaluated using exact-match accuracy. Full dataset statistics are in Appendix~\ref{tab:datasets}.

\begin{table*}[ht!]
\centering
\small
\setlength{\tabcolsep}{5pt}
\renewcommand{\arraystretch}{1.15}
\resizebox{\textwidth}{!}{%
\begin{tabular}{>{\raggedright\arraybackslash}m{2.5cm}
                l
                c c c c }
\specialrule{1pt}{0pt}{1.5pt}
 & & \multicolumn{1}{c}{\textbf{In-Domain}} & \multicolumn{3}{c}{\textbf{Out-of-Domain (OOD)}} \\
\cmidrule{3-3} \cmidrule{4-6}
\multicolumn{1}{c}{\textbf{Category}} & \textbf{Model} & \textbf{MedAraBench} & \textbf{MedArabiQ} & \textbf{AraSTEM} & \textbf{ArabicMMLU} \\
\specialrule{1pt}{0pt}{1.5pt}
\multicolumn{1}{c}{Random} & Random Baseline & 24.0 & 20.0 & 21.1 & 29.7  \\
\specialrule{0.6pt}{0pt}{0pt}
\multirow{3}{3.0cm}{\parbox[c]{3.0cm}{\centering Closed-Source\\General-Purpose}}
& GPT-5.2~\citep{openai2025gpt52} & 71.0 & 78.9 & 88.0 & 71.6  \\
& Gemini-2.5-Flash~\citep{comanici2025gemini25pushingfrontier} & \underline{70.2} & \underline{79.0} & \underline{85.2} & \underline{73.8}  \\
& Claude-Opus-4.6~\citep{anthropic2026claudeopus46} & \textbf{74.3} & \textbf{82.1} & \textbf{89.4} & \textbf{74.1}  \\
\specialrule{0.6pt}{0pt}{0pt}
\multirow{6}{3.0cm}{\parbox[c]{3.0cm}{\centering Open-Source\\General-Purpose}}
& Mistral-7B-Instruct-v0.3~\citep{jiang2023mistral7b} & 27.2 & 25.2 & 26.6 & 31.3  \\
& Llama-3.1-8B-Instruct~\citep{grattafiori2024llama3herdmodels} & 37.7 & 32.6 & 37.8 & 38.6  \\
& Mistral-Small-3.2-24B~\citep{mistral2025small32} & \underline{52.6} & \underline{51.5} & 62.6 & \underline{55.2}  \\
& Gemma-3-27B-IT~\citep{gemmateam2025gemma3technicalreport} & 51.6 & 47.3 & \underline{64.2} & 53.8  \\
& Llama-3.3-70B-Instruct~\citep{grattafiori2024llama3herdmodels} & 46.8 & 40.0 & 55.0 & 51.0  \\
& DeepSeek-V3.2~\citep{deepseekai2025deepseekv32pushingfrontieropen} & \textbf{63.4} & \textbf{66.3} & \textbf{76.9} & \textbf{66.0}  \\
\specialrule{0.6pt}{0pt}{0pt}
\multirow{6}{3.0cm}{\parbox[c]{3.0cm}{\centering Arabic/Multilingual\\General-Purpose}}
& Jais-2-8B-Chat~\citep{anwar2026jais} & 37.9 & 30.5 & 40.9 & \textbf{49.3}  \\
& ALLaM-7B-Instruct-Preview~\citep{bari2025allam} & \underline{43.8} & \underline{36.8} & \underline{46.1} & 46.2  \\
& Aya-Expanse-8B~\citep{dang2024ayaexpansecombiningresearch} & 39.6 & 34.7 & 43.1 & 42.6 \\
& SILMA-9B-Instruct-v1.0~\citep{silmaai2026silma9b} & 42.2 & 34.0 & 42.8 & 43.9  \\
& Falcon-H1-7B-Instruct~\citep{zuo2025falconh1familyhybridheadlanguage} & \textbf{46.7} & \textbf{40.0} & \textbf{48.6} & 45.6 \\
& Fanar-1-9B-Instruct~\citep{fanarteam2025fanararabiccentricmultimodalgenerative} & 40.8 & 33.6 & 44.3 & \underline{47.0}  \\
\specialrule{0.6pt}{0pt}{0pt}
\multirow{3}{3.0cm}{\parbox[c]{3.0cm}{\centering Medical-Domain}}
& MedGemma-27B-Text-IT~\citep{sellergren2026medgemmatechnicalreport} & \underline{51.9} & \textbf{51.5} & \underline{63.8} & 51.3  \\
& Meditron-3-70B~\citep{sallinen2025llama} & \textbf{55.9} & \underline{50.5} & \textbf{66.4} & \textbf{55.9}  \\
& Llama-3-Med42-70B~\citep{christophe2024med42v2suiteclinicalllms} & 39.0 & 29.0 & 62.8 & \underline{51.9}  \\

\specialrule{0.6pt}{0pt}{0pt}
\multirow{8}{3.0cm}{\parbox[c]{3.0cm}{\centering Adaptation\\Methods}}
& Mistral + Few-Shot (k=5)~\citep{brown2020languagemodelsfewshotlearners} & 53.7 & 46.3 & 61.5 & 57.6  \\
& Mistral + AUTOCAP~\citep{zhang-etal-2024-autocap} & 59.7 & 54.0 & \underline{68.4} & 59.6 \\
& Mistral + English Translation & 57.8 & 54.0 & \textbf{71.5} & 56.8 \\
& Mistral + MindMerger~\citep{huang2024mindmerger} & 24.0 & 22.0 & 20.3 & 25.1  \\
& BiMedix (Zero-shot)~\citep{pieri-etal-2024-bimedix} & 28.6 & 28.0 & 37.5 & 29.3  \\
& Mistral + LoRA ~\citep{hu2022lora} & 42.6 & 42.0 & 45.5 & 36.3 \\
& Mistral + LoRA v2 (with KL)& \underline{61.9} & \underline{55.0} & 60.7 & \underline{60.3} \\
\rowcolor{gray!25}& Mistral + TLoRA (Ours) & \textbf{62.1} & \textbf{60.0} & 65.1 & \textbf{61.6}  \\
\specialrule{1pt}{0pt}{0pt}

\end{tabular}%
}
\caption{Multiple-Choice QA results across four Arabic medical benchmarks. Exact-match accuracy (\%) is reported on MedAraBench (in-domain) and MedArabiQ, AraSTEM-medicine, and ArabicMMLU-biology (out-of-domain benchmarks). Bold indicates the best result within each group per column, and underline points to second best result. The Mistral variant used as backbone for the adaptation methods is Mistral-Small-3.2-24B-Instruct-2506.}
\label{tab:mcq-results}
\end{table*}

\subsubsection{Short Answer Generation}
\label{sec:short-answer}

We repurpose MedAraBench for free-text generation. Starting from $4,989$ test examples, we remove MCQA artifacts (e.g., ``All of the above'') and rewrite remaining questions into standalone open-ended queries via an LLM-assisted pipeline, yielding 940 examples. We refer to this subset 
as \textbf{MedAraBench-OE} (Open-Ended). Full preprocessing details are in Appendix~\ref{app:data-preprocessing}.

Standard n-gram metrics are poorly suited for Arabic medical generation due to morphological variability and short reference answers. We therefore use LLM-as-a-judge (GPT-5.2) as our primary metric, validated against human judgements on a 100-example subset: LLM-judge accuracy achieves Pearson $= 0.978$ and Spearman $= 0.982$ with human labels, substantially 
outperforming BERTScore-F1 (Pearson $= 0.802$, Spearman $= 0.715$). BERTScore-F1 (AraBERTv2) is reported as a secondary metric. Full reliability analysis is in Appendix~\ref{app:llm-judge-reliability}.

\subsubsection{Multi-Turn Clinical Dialogue}

We introduce \textbf{AraClinicDialog}, constructed by three native Arabic-speaking physicians from a multi-specialty hospital. Following \citet{arora2025healthbenchevaluatinglargelanguage}, physicians authored 100 clinical cases spanning nine organ systems, expanded into multi-turn dialogues using Claude-Opus-4.6 and verified by clinicians. Each physician authored an independent reference response
for the final turn, while a fourth clinician assessed agreement, achieving 87\% inter-annotator agreement. Full construction details are in Appendix~\ref{app:task3-data-collection}. 

Dataset construction proceeded in four stages:
(a) Case template authoring: each physician wrote a clinical scenario spanning nine organ systems, yielding 100 cases (Appendix~\ref{tab:organ-systems}).
(b) Dialogue generation: each template was expanded into a
3--8 turn MSA patient-assistant dialogue using Claude-Opus-4.6, with
the final assistant turn withheld (Appendix~\ref{fig:dialogue-gen-prompt}). (c) Reference collection: two clinicians per case
independently authored the withheld final turn via a structured form. Audio responses were transcribed with Whisper-Large-V3 and manually reviewed (Appendix~\ref{tab:reference-answers}). (d) Dialect translation: all 100 MSA dialogues were translated into four regional Arabic dialects, Emirati, Jordanian, Moroccan and Egyptian, using GPT-5.2 and reviewed by two native speakers per dialect (Appendix~\ref{tab:dialect-example}).

Models are evaluated on the final dialogue turn using correct/incorrect judgement against the clinician-written reference, anchored in a primary reasoning objective authored by clinicians, and assessed by LLM-as-a-judge. This framing is consistent with Task~2
and targets factual correctness of the concluding clinical response
rather than dialogue quality overall. Given the validated alignment
between LLM-judge and human labels established in
\S\ref{sec:short-answer}, we use the same metrics.

\begin{figure*}[ht!]
    \centering
        \includegraphics[width=\textwidth]{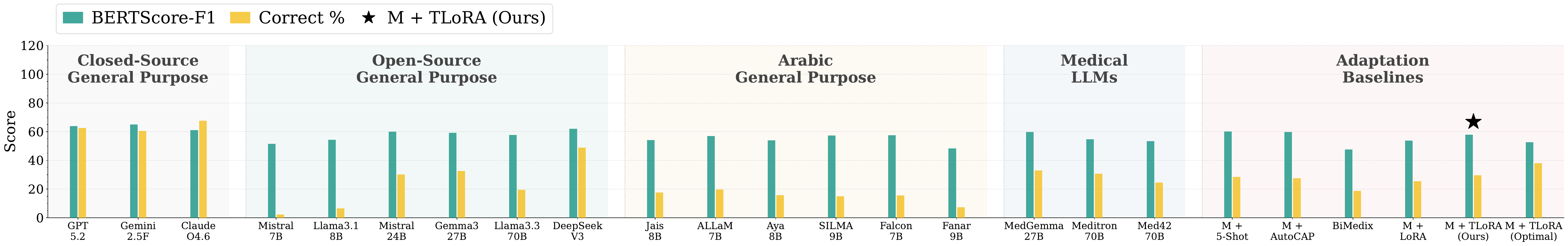}
    \caption{Short answer generation performance across model categories, evaluated on MedArabench-OE. For each model, we report BERTScore-F1 (araBERTv2) and LLM-as-a-judge Correct
        \%. The star ($\star$) denotes our proposed method.}
    \label{fig:task2-results}
\end{figure*}

\begin{figure*}[ht!]
    \centering
        \includegraphics[width=\textwidth]{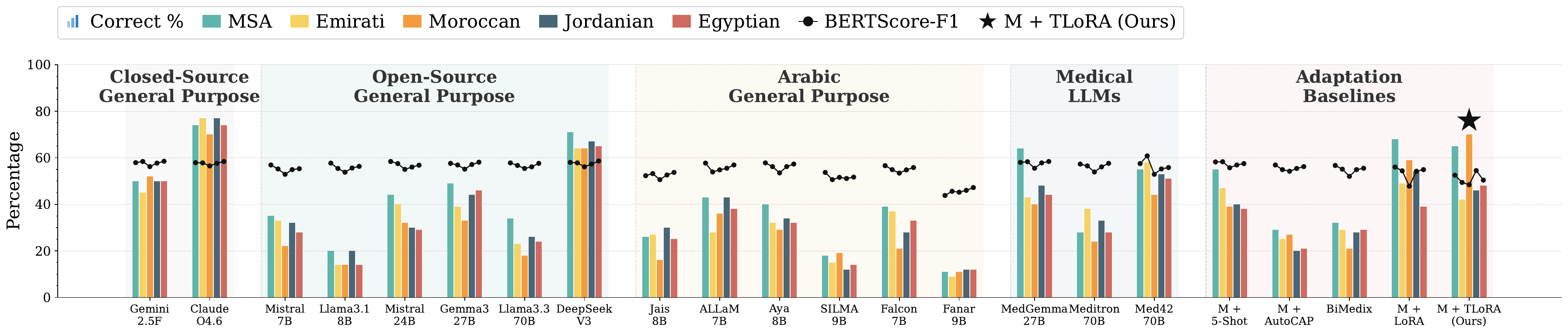}
        \caption{Multi-Turn Clinical Dialogue performance across model categories, evaluated on AraClinicDialog (MSA and dialect variants). For each model, we report BERTScore-F1 (AraBERTv2) and LLM-as-a-judge Correct
        \%. The star ($\star$) denotes our proposed method.}
    \label{fig:task3_bertscores_avg}
\end{figure*}

\subsection{Baselines}
Adaptation baselines include few-shot prompting ($k=5$) ~\citep{brown2020languagemodelsfewshotlearners}, AUTOCAP ~\citep{zhang-etal-2024-autocap}, purpose-built Arabic medical model BiMedix ~\citep{pieri-etal-2024-bimedix}, full LoRA ~\citep{hu2022lora}, and MindMerger ~\citep{huang2024mindmerger}. The latter two are trained on the MedAraBench training split, while the remaining baselines are prompt-based. MindMerger is evaluated on MCQA only, as it struggled to produce the required output format on generation and dialogue tasks.

\section{Results}

\subsection{MCQA}

Table~\ref{tab:mcq-results} shows results across the four Arabic medical MCQA benchmarks. Closed-source models form a clear upper tier, with macro-averages between 77.1\% and 80.0\%. Among open-source general-purpose models, performance varies substantially: Mistral-Small-3.2-24B (55.5\%) and DeepSeek-V3.2 (68.2\%) substantially outperform smaller models, though scale alone is not predictive: Llama-3.3-70B underperforms Mistral-Small-3.2-24B despite its larger size. Notably, Arabic/multilingual models do not consistently outperform general open-source models, averaging in the low-to-mid 40\% range; Arabic-specific pretraining alone does not resolve the knowledge-access gap, consistent with prior work \citep{pmlr-v298-daoud25a}.

Among adaptation methods, MindMerger degrades substantially below the zero-shot Mistral baseline, and few-shot prompting offers no reliable gain. LoRA v2 improves over zero-shot (59.5\% vs.\ 55.5\%) but applies adaptation uniformly across all 40 layers. Our method, TLoRA, achieves the highest macro-average among adaptation methods (62.2\%), outperforming LoRA (w/o KL) by 20.6 points and LoRA v2 (with KL) by 2.7 points despite training fewer parameters. TLoRA and LoRA v2 converge to a similar ceiling on MedAraBench
(in-domain): 95\% confidence intervals overlap substantially (TLoRA [60.8, 63.4] vs.\ LoRA v2 [60.5, 63.2], $p=0.74$). The out-of-domain comparisons are more diagnostic, where TLoRA's advantage is clearest on MedArabiQ and ArabicMMLU. This advantage holds under controlled conditions: fixing the learning rate to the same value across all windows and removing the KL alignment term both preserve the L1--34 ranking, ruling out optimisation and loss formulation as confounds (Appendix~\ref{app:layer-vs-kl}, \ref{app:fixed-lr}). This advantage holds under controlled conditions: fixing the learning rate to the same value across all windows and removing the KL alignment term both preserve the L1--34 ranking, ruling out optimisation and loss formulation as confounds (Appendix~\ref{app:layer-vs-kl}, \ref{app:fixed-lr}).

\begin{table*}[!t]
\centering
\small
\setlength{\tabcolsep}{2pt}
\renewcommand{\arraystretch}{1.15}
\resizebox{\textwidth}{!}{%
\begin{tabular}{l cccc cc cc}
\specialrule{1pt}{0pt}{1.5pt}
& \multicolumn{4}{c}{\textbf{MCQA}} 
& \multicolumn{2}{c}{\textbf{Gen.}} & \multicolumn{2}{c}{\textbf{Dial.}} \\
\cmidrule(lr){2-5}\cmidrule(lr){6-7}\cmidrule(lr){8-9}
\textbf{Window} 
  & \textbf{MedAraBench} & \textbf{MedarabiQ} 
  & \textbf{MMLU-Bio} & \textbf{AraSTEM} 
  & \textbf{BERTScore} & \textbf{LLM Judge} 
  & \textbf{BERTScore} & \textbf{LLM Judge} \\
\specialrule{1pt}{0pt}{1.5pt}
\rowcolor{gray!25}
Zero-shot                          & 52.6 & 51.5 & 55.2 & 62.6 & 60.0 & 30.1 & 58.4 & 44.0 \\
Full LoRA ($W_5$: L1--40)          & \underline{61.9} & 55.0 & \underline{60.3} & 60.7 & \underline{53.8} & 25.5 & \underline{56.0} & 68.0 \\
\specialrule{0.6pt}{0pt}{0pt}
Targeted ($W_1$: L1--24)           & 60.7 & \underline{56.0} & 60.2 & \underline{63.2} & 53.4 & 29.0 & \textbf{57.2} & 72.0 \\
Targeted ($W_2$: L24--40)          & 53.5 & 42.0 & 52.9 & 58.5 & 52.6 & \textbf{38.1} & 48.3 & \textbf{76.0} \\
Targeted ($W_3$: L1--34)           & \textbf{62.1} & \textbf{60.0} & \textbf{61.6} & \textbf{65.1} & \textbf{57.9} & \underline{29.7} & 52.5 & 64.0 \\
Targeted ($W_4$: L34--40)          & 47.3 & 33.0 & 51.3 & 54.6 & 50.6 & 26.5 & 48.3 & \underline{75.0} \\
\specialrule{1pt}{0pt}{0pt}
\end{tabular}%
}
\caption{Window ablation. All targeted variants use $\mathcal{L}_\text{CE} + \beta^*\mathcal{L}_\text{align}$ with probe layer $L_\text{KL}=34$ and bilingual training data. Short Answer Generation (Gen.) and Multi-Turn Clinical Dialogue (Dial.) scores are BERTScore-F1 and LLM-judge accuracy (\%). Bold indicates the best result among trained variants for each column, while underline indicates the second-best result.}
\label{tab:window_ablation}
\end{table*}

\subsection{Short Answer Generation}

Figure~\ref{fig:task2-results} reports results on MedAraBench-OE, a generalization task no adaptation method was optimized for. Closed-source models lead, with LLM-judge accuracy between 60.5\% and 67.6\%; smaller open-source and Arabic/multilingual models largely fail to produce coherent free-text responses. Among adaptation methods, MCQA fine-tuning generally degrades generation performance relative to zero-shot: Full LoRA drops from 30.1\% to 25.5\%. Our method retains 29.7\% [26.8\%, 32.6\%], statistically indistinguishable from zero-shot (30.1\% [27.2\%, 33.0\%], $p=0.86$), and the closest to zero-shot among trained variants, suggesting that targeted adaptation does not induce the generation-forgetting observed in broader fine-tuning. 

TLoRA is selected by MCQA performance while TLoRA (optimal), also shown in Figure~\ref{fig:task2-results}, instead uses the window best suited to this task and reaches higher generation accuracy (38.1\%). Ablation results indicate that the KL alignment term contributes specifically to this preservation: removing it while keeping the same window reduces generation BERTScore substantially, whereas MCQA accuracy is largely unchanged (Appendix~Table \ref{tab:ce-only-windows}). Complete results for Short Answer Generation on MedAraBench-OE are provided in Appendix~\ref{app:complete-results-task2}.

\subsection{Multi-Turn Clinical Dialogue}


Figure~\ref{fig:task3_bertscores_avg} reports results on AraClinicDialog's MSA and dialect variants. Unlike Short Answer Generation, adaptation yields substantial gains on dialogue over zero-shot. Both full LoRA and TLoRA improve, with our method remaining competitive despite training exclusively on MCQA data. AUTOCAP, strong on MCQA, collapses on dialogue, suggesting its gains are format-sensitive. Closed-source models remain the upper bound (complete per-dialect results are reported in Appendix~\ref{app:complete-results-task3}, Tables~\ref{tab:task3-msa-results}--\ref{tab:task3-egyptian-results}).

\subsection{Ablations}

Table~\ref{tab:window_ablation} reports performance across all five candidate adaptation windows. On MCQA, L1--34 achieves the highest macro-average, outperforming full-network LoRA and all other windows. The result supports the mechanistic hypothesis: restricting adaptation to layers below the divergence onset outperforms both broader and narrower windows. The L1--34 window is selected based on the mechanistic diagnosis on MCQA, rather than downstream performance; other windows may therefore outperform it on individual downstream metrics. On dialogue, no trained variant falls below zero-shot. The L1--34 advantage is consistent across loss functions and learning-rate conditions. Detailed comparisons
against CE-only and fixed-LR baselines are provided in Appendix~\ref{app:layer-vs-kl}, \ref{app:fixed-lr}. Sensitivity of results to the KL probe layer threshold is reported in Appendix~\ref{sec:kl-sensitivity}. We adopt L1--34 as our method for all subsequent comparisons.

\section{Discussion}


\paragraph{MCQA.}
The performance ordering across windows is directionally consistent with the mechanistic hypothesis: gains scale with the degree to which the adaptation window covers the identified routing failure region, rather than with the number of layers adapted. CE-only training across all windows (Appendix~\ref{app:layer-vs-kl}) shows that L1--34 retains its MCQA lead without $\mathcal{L}_\text{align}$, confirming that window placement is the primary driver, while the alignment term contributes selectively to generation quality rather than classification accuracy.  TLoRA also narrows the access gap identified in Section~\ref{sec:mechanistic}, reducing the English-correct/Arabic-incorrect failure rate from 29.6\% (zero-shot) to 19.0\%, slightly ahead of Full LoRA's 20.2\%. The same diagnostic pipeline applied to a second model family, Llama-3.1-8B-Instruct, identifies a different window that matches or exceeds full-network LoRA on MCQA (Appendix~\ref{app:generalizability}), suggesting the approach is not specific to Mistral.

MindMerger's degradation below the zero-shot baseline is architecturally informative: having been fine-tuned on bilingual medical terminology prior to MedAraBench training, domain mismatch can be excluded as an explanation. Encoder augmentation with a frozen backbone intervenes at the input representation level, leaving the late-layer routing deficit entirely unaddressed. Few-shot prompting is similarly ineffective for the same underlying reason: surface-level context cannot recover knowledge that is representationally accessible but fails to route to the output in Arabic.

\paragraph{Generation and Dialogue.}
MCQA fine-tuning broadly degrades generation, likely through format overfitting; our method is least affected, suggesting targeted adaptation preserves more general capability. On dialogue, the pattern is reversed: adaptation helps substantially. The high dialogue scores are partly explained by the evaluation design: the judge is anchored in clinician-authored reasoning objective field (Appendix Figure~\ref{tab:template-example-1}), rewarding clinical accuracy over fluency. Manual inspection confirms responses were short and frequently code-switched, performing well factually while likely underperforming on communicative quality.

\section{Conclusion and Future Work}
We present TLoRA, a mechanistically-grounded approach to Arabic medical adaptation: diagnosing where a model fails directly informs where adaptation should intervene. Tuned lens probing and causal activation patching identify a knowledge-routing failure localised to a specific layer window. Restricting LoRA to that window outperforms full-network adaptation and other baselines. Results on generation and dialogue further show that targeted adaptation preserves general capability where broader fine-tuning does not, offering a tractable path toward closing the Arabic medical NLP gap.

Future work will test whether the observed routing failure generalises to other model families and architectures. We also plan to scale AraClinicDialog to further dialects and clinical scenarios. Whether targeted adaptation extends to other low-resource languages with similar knowledge-access gaps remains to be tested.

\FloatBarrier
\section*{Limitations}
First, our mechanistic analysis and proposed TLoRA method are developed on the Mistral-Small-3.2-24B model family and further validated on Llama-3.1-8B-Instruct. Whether the observed localized routing failure generalises to other decoder-only models, encoder-decoder architectures, or larger-scale checkpoints remains untested.

Second, AraClinicDialog is limited in scale, with only 500 dialogues currently included, which constrains its coverage of rare and low-frequency medical scenarios. Additionally, while the benchmark covers Modern Standard Arabic and four major dialects, it does not yet encompass all regional Arabic varieties.


\section*{Ethical Considerations}
This study investigates cross-lingual knowledge routing and mechanistic adaptation methods for improving Arabic medical language understanding in LLMs. The AraClinicDialog benchmark was constructed with the involvement of clinician co-authors, who authored hypothetical clinical case scenarios and reference dialogue responses; these scenarios do not correspond to real patients, and the research does not involve the use of proprietary, private, or sensitive patient data. All datasets, pretrained models, and external resources used in this work comply with their respective licenses and terms of use. The proposed methodology is intended to improve access to medical knowledge in under-represented languages such as Arabic and can contribute to more equitable multilingual medical NLP systems. It is not intended as a substitute for clinicians, medical advice, diagnosis, or treatment. Future work should further evaluate the robustness, safety, and generalizability of these methods across broader medical settings and languages.

\section*{Acknowledgements}
This work was supported by the Meem Foundation and the New York University Abu Dhabi (NYUAD) Center for Interdisciplinary Data Science and AI (CIDSAI), funded by Tamkeen under the NYUAD Research Institute Award CG016. The research was carried out on NYUAD’s High Performance Computing resources (Jubail).

\bibliography{custom}

\appendix
\setcounter{figure}{0}
\setcounter{table}{0}

\renewcommand{\thefigure}{S\arabic{figure}}
\renewcommand{\thetable}{S\arabic{table}}
\newpage


\section{Experimental Setup} 
\label{app:exp-setup}

\subsection{Datasets}

Table \ref{tab:datasets} summarizes the six benchmarks used in our evaluation, spanning multiple-choice QA, short-answer generation, and multi-turn dialogue tasks. AraClinicDialog is introduced in this work, MedAraBench-OE repurposes MedAraBench's existing questions into an open-ended format, and the remaining four are existing benchmarks used as-is.

\begin{table*}[t]
\centering
\small
\begin{tabular}{llrl}
\toprule
\textbf{Dataset} & \textbf{Task} & \textbf{Size} & \textbf{Domain} \\
\midrule
MedAraBench \citep{daoud2026medarabench}      & MCQA                & 4,959 & Clinical medicine \\
MedArabiQ \citep{pmlr-v298-daoud25a}        & MCQA                &   100 & Clinical medicine \\
ArabicMMLU \cite{koto-etal-2024-arabicmmlu}     & MCQA                & 1,072 & Medicine \& biology \\
AraSTEM  \cite{boussaha-etal-2025-3lm}        & MCQA                &   721 & STEM / medicine \\
MedAraBench-OE   & Short Answer Generation  &    940 & Clinical medicine \\
AraClinicDialog  & Multi-Turn Dialogue &    100 & Clinical medicine \\
\bottomrule
\end{tabular}
\caption{Dataset statistics for all evaluation benchmarks.}
\label{tab:datasets}
\end{table*}

\subsection{Prompts}
\label{app:prompts}

Prompts are defined per task. For the MCQA task, all baselines share a single zero-shot prompt. For the Short Answer Generation and the Multi-Turn Clinical Dialogue tasks, we use a one-shot setup to ensure models adhere to the required answer format. The few-shot baselines are the exception, which use few-shot prompt variants. 

The Short Answer Generation and the Multi-Turn Clinical Dialogue tasks are evaluated using an LLM-as-a-judge protocol, in addition to automatic metrics such as BERTScore. The corresponding judge prompts are provided in the relevant sections.

\textbf{MCQA. } The zero-shot prompt is shown in Figure~\ref{fig:prompt-task1-zero}. The few-shot variant extends it with five exemplar question-answer pairs. The exemplars are dataset-specific, randomly sampled with a fixed seed (42), and excluded from the inference set. The diagnostic mechanistic-analysis pass in \S\ref{sec:mechanistic} uses a separate, dedicated prompt, shown in Figure~\ref{fig:prompt-mech}.

\begin{figure}[ht]
\begin{tcolorbox}[
  colback=boxbg,
  colframe=headernavy,
  coltitle=white,
  fonttitle=\normalsize\bfseries,
  title=Zero-shot Prompt (MCQA task),
  arc=4pt,
  boxrule=1pt,
  left=6pt, right=6pt,
  top=4pt, bottom=4pt,
  fontupper=\normalsize,
]
\setlength{\parskip}{5.5pt}
\setlength{\parindent}{0pt}

You are a medical expert answering multiple-choice exam questions. \\ \\
You will receive exactly ONE question followed by answer options labeled: \\
A), B), C), D), E), and sometimes F). \\ \\
You must output exactly ONE line in this format: \\
ANSWER: <LETTER> \\ \\
Rules: \\
- Output ONLY that line. \\
- Do NOT repeat or paraphrase the question. \\
- Do NOT translate anything. \\ 
- Do NOT explain your reasoning. \\ 
- Do NOT list the options.
\end{tcolorbox}
\caption{Zero-shot prompt for the MCQA task.}
\label{fig:prompt-mech}
\end{figure}

\begin{figure}[ht]
\begin{tcolorbox}[
  colback=boxbg,
  colframe=headernavy,
  coltitle=white,
  fonttitle=\normalsize\bfseries,
  title=Zero-shot Prompt (Mechanistic Analysis),
  arc=4pt,
  boxrule=1pt,
  left=6pt, right=6pt,
  top=4pt, bottom=4pt,
  fontupper=\normalsize,
]
\setlength{\parskip}{5.5pt}
\setlength{\parindent}{0pt}
You are a medical assistant specialized in solving multiple-choice exam questions.\\
You will receive exactly ONE question, followed by answer options labeled with capital Latin letters:\\
A), B), C), D), E) and sometimes F).\\

Your task:\\
1. Read the question and all options carefully.\\
2. Decide which single option (A–F) is MOST correct.\\
3. DO NOT translate the question or explain your reasoning.\\

Output format (VERY IMPORTANT):\\
- Your final answer MUST be on a single line in this exact format:\\

ANSWER: <LETTER> \\where <LETTER> is exactly one of: A, B, C, D, E, F.\\

- Output ONLY the letter after "ANSWER:".\\
- Do NOT output the option text.\\
- Do NOT add any extra words, explanations, punctuation, or symbols.\\
\end{tcolorbox}
\caption{Zero-shot prompt for the Mechanistic Analysis.}
\label{fig:prompt-task1-zero}
\end{figure}





\textbf{Short Answer Generation. }  The prompt is shown in Figure~\ref{fig:prompt-task2}. We use a one-shot rather than a zero-shot setup because, under zero-shot, most models failed to follow the short-answer instruction and produced verbose explanatory responses; a single in-context example was sufficient to enforce the intended output format. The example therefore calibrates both answer style and length. 

As in the MCQA task, we define a separate few-shot prompt variant for that setting: the base prompt is extended with five exemplar question-answer pairs, which are randomly sampled using a fixed seed (42) and are excluded from the inference set.

For LLM-as-a-judge evaluation, the judge prompt (Figure~\ref{fig:prompt-task2-judge}) takes the question, reference answer, and generated answer as input and returns a binary correctness label.

\begin{figure}[!t]
\begin{tcolorbox}[
  colback=boxbg,
  colframe=headernavy,
  coltitle=white,
  fonttitle=\normalsize\bfseries,
  title=One-shot Prompt (Short Answer Generation),
  arc=4pt,
  boxrule=1pt,
  left=6pt, right=6pt,
  top=4pt, bottom=4pt,
  fontupper=\normalsize,
]
\setlength{\parskip}{5.5pt}
\setlength{\parindent}{0pt}

You are a medical expert.\\
You will receive exactly ONE medical exam question with no answer options.
Answer the question based on your medical knowledge and return your answer as a short medical term or phrase.\\ \\
Output format:\\
ANSWER: <short answer>\\ \\
Rules:\\
- Output exactly ONE line starting with “ANSWER: “.\\
- The answer must be a concise medical term or phrase, typically 1-4 words, rarely more than 10 words.\\
- All answers must be in Arabic.\\
- Do NOT write a full sentence or explanation.\\
- Do NOT explain your reasoning.\\
- Do NOT add extra commentary.\\
- Do NOT repeat the question.\\
- Do NOT translate.\\
- Do NOT include multiple answers. \\ \\
Example: \\
Question: \textarabic{أين تتوضع الثقبة العوراء في اللسان؟} \\
ANSWER: \textarabic{خلف الكلم الإنتهائي}




\end{tcolorbox}
\caption{One-shot prompt for the Short Answer Generation.}
\label{fig:prompt-task2}
\end{figure}

\begin{figure}[!htbp]
\begin{tcolorbox}[
  colback=boxbg,
  colframe=headernavy,
  coltitle=white,
  fonttitle=\normalsize\bfseries,
  title=LLM-as-a-Judge Prompt (Short Answer Generation task),
  arc=4pt,
  boxrule=1pt,
  left=6pt, right=6pt,
  top=4pt, bottom=4pt,
  fontupper=\normalsize,
]
\setlength{\parskip}{5.5pt}
\setlength{\parindent}{0pt}
You are an expert medical evaluator. You will be given a medical question, a reference answer, and a generated answer. Your task is to evaluate the generated answer by selecting exactly one label from the following options and responding only with the label in brackets []. \\ 
Question: \{question\_stem\} \\
Reference Answer: \{reference\_answer\} \\
Generated Answer: \{generated\_answer\} \\ \\
Evaluate the generated answer against the reference answer using the following criterion: \\ \\
Correctness: Correct / Incorrect \\ \\
- Correct: The generated answer matches the reference answer in meaning. Synonyms, equivalent medical terms, or slight phrasing differences still count as correct. \\
- Incorrect: The generated answer is wrong, irrelevant, contradicts the reference answer, or is too vague to be credited. \\ \\
Respond only with one of the following in brackets: [Correct] / [Incorrect]
\end{tcolorbox}
\caption{LLM-as-a-judge prompt for the Short Answer Generation task.}
\label{fig:prompt-task2-judge}
\end{figure}

\textbf{Multi-Turn Clinical Dialogue. }  The prompt is shown in Figure~\ref{fig:prompt-task3}. Similar to the Short Answer Generation task, we use a one-shot setup due to models struggling to follow the expected output format under zero-shot.

For the few shot baseline, we define a separate few-shot prompt variant for that setting: the base prompt is extended with five carefully constructed question-answer pairs to encourage the model to produce responses that reflect the intended reasoning approach and structure.

For the LLM-as-a-judge evaluation metric, the judge prompt (Figure~\ref{fig:prompt-task3-judge}) takes the primary reasoning objective, red-flag symptoms, dialogue, and generated answer as input and assigns one of three labels along each of three axes: Reasoning Match, Safety, and Communication. This design reflects the multi-dimensional nature of the task, which requires not only clinically appropriate reasoning but also safe medical guidance and effective communication.

\begin{figure}[!htbp]
\begin{tcolorbox}[
  colback=boxbg,
  colframe=headernavy,
  coltitle=white,
  fonttitle=\normalsize\bfseries,
  title=One-Shot Prompt (Multi-Turn Clinical Dialogue),
  arc=4pt,
  boxrule=1pt,
  left=6pt, right=6pt,
  top=4pt, bottom=4pt,
  fontupper=\normalsize,
]
\setlength{\parskip}{5.5pt}
\setlength{\parindent}{0pt}
You are an expert medical doctor. You will be given a conversation between a doctor and a patient. The final doctor response is missing. Based on the conversation, generate the doctor's concluding response, typically a diagnosis, recommendation, or treatment plan. \\\\
Output format:\\
ANSWER: <doctor's response>\\\\
Rules:\\
- Output exactly ONE line starting with "ANSWER: ".\\
- The response must be in Modern Standard Arabic. \\
- Be concise and clinically appropriate (1–3 sentences maximum).\\
- Do NOT repeat or summarize the conversation.\\
- Do NOT add extra commentary or explanation. \\
- Do NOT translate anything.\\\\
Example:\\\\
\{Sample Arabic Clinical Dialogue Between Doctor and Patient\} \\\\
ANSWER: \textarabic{هذه العوارض هي كوشينج سندروم وتعتبر مرض خطير إذا لم يتم علاجه أولا. عليكي أن تراجعي طبيب مختص في الغدد في أسرع وقت، وعمل إجراءات فحوص يتضمن مقطعية وفحص دم. بس لا تقلقي هذا المرض له علاج وممكن أن تشفي منه تماما.}

\end{tcolorbox}
\caption{One-shot prompt for the Multi-Turn Clinical Dialogue task. The full dialogue is omitted for space and replaced with a placeholder. We retain the example answer to illustrate the expected output format. The complete dialogue sample is shown in Figure~\ref{fig:dialogue-example-1}.}
\label{fig:prompt-task3}
\end{figure}

\begin{figure*}[!t]
\begin{tcolorbox}[
  colback=boxbg,
  colframe=headernavy,
  coltitle=white,
  fonttitle=\normalsize\bfseries,
  title=LLM-as-a-Judge Prompt (Multi-Turn Clinical Dialogue),
  arc=4pt,
  boxrule=1pt,
  left=6pt, right=6pt,
  top=4pt, bottom=4pt,
  fontupper=\normalsize,
]
\setlength{\parskip}{5.5pt}
\setlength{\parindent}{0pt}
You are an expert medical evaluator. You will be given a doctor-patient dialogue (in Arabic), the Primary Reasoning Objective that the final doctor turn was supposed to clinically reach, and a generated final doctor turn. Your task is to evaluate whether the generated turn correctly achieves the Primary Reasoning Objective, and respond only with the label in brackets [].\\\\
Dialogue: \{dialogue\}\\\\
Primary Reasoning Objective: \{primary\_reasoning\_objective\}\\\\
Generated Answer: \{generated\_answer\}\\\\
Evaluate the generated answer against the Primary Reasoning Objective using the following criterion:\\\\
Correctness: Correct / Incorrect\\\\
- Correct: The generated answer reaches or is consistent with the diagnosis, differential, or management direction in the Primary Reasoning Objective. This includes answers that are in the right clinical direction even if they do not fully articulate every detail. Synonyms, equivalent medical terms, and slight phrasing differences still count as correct. Do not penalize for code-switching or English medical terminology if the clinical content is correct.\\\\
- Incorrect: The generated answer fails to engage with the Primary Reasoning Objective, identifies a clearly different diagnosis, recommends a contradictory course of action, or is too vague to demonstrate any clinical reasoning.\\\\
Respond only with one of the following in brackets: [Correct] / [Incorrect]
\end{tcolorbox}
\caption{LLM-as-a-judge prompt for the Multi-Turn Clinical Dialogue task.}
\label{fig:prompt-task3-judge}
\end{figure*}

\FloatBarrier
\section{Data Collection and Pre-processing}

This appendix specifies the data pre-processing pipeline for the Short Answer Generation task (Section~\ref{app:data-preprocessing}) and the dataset construction procedure for the Multi-Turn Clinical Dialogue task (Section~\ref{app:task3-data-collection}).

\subsection{Adapting MCQA Data for Open-Ended Answer Generation}
\label{app:data-preprocessing}




To evaluate free-form medical answer generation in Arabic, we needed a dataset that differs from MCQA while still testing medical knowledge. Two natural candidates exist: AraMed \citep{alasmari-etal-2024-aramed}, which is the only native Arabic QA benchmark to the best of our knowledge, and English medical QA benchmarks translated into Arabic. Both have limitations. AraMed is sourced from public medical forum Altibbi, creating a high risk of contamination. Translation, in turn, incurs well-documented information loss, particularly for specialized medical terminology, and conflicts with our broader commitment to evaluating models exclusively on native Arabic benchmarks. 

We therefore repurpose MedAraBench, the largest MCQA dataset used in our MCQA task, into an open-ended generation benchmark through a four-stage pipeline: automated screening, manual verification, automated question reformulation, and a final manual quality pass. The final repurposed benchmark is referred to as MedAraBench-OpenEnded (MedAraBench-OE hereafter).

\subsubsection{Automated Screening}
\label{app:screening}
 
The source dataset comprised 4,958 multiple-choice questions. To filter out items structurally incompatible with conversion to an open-ended format, we prompted  GPT‑5.2 to label each sample as \textbf{yes} (keep), \textbf{no} (drop), or \textbf{maybe} (borderline). The classification criteria were grounded in the structural requirements of open-ended answer generation: a valid open-ended question must elicit a direct, standalone response without relying on a predefined set of answer choices. 
 
Questions were flagged for removal if they fell into any of the following categories: 
\begin{itemize}
    \item Exclusion-logic questions using phrasing, which test the ability to
identify a single false item among otherwise correct options
    \item Questions with options that cross-reference one another
    \item Questions that ask the test-taker to identify the incorrect or
false statement rather than the correct one
    \item Items that are not properly formed as questions, such as entries that merely label a body part or anatomical structure
    \item Items in which the question stem or options are primarily in English
\end{itemize}

We prompt GPT-5.2 in a few-shot setting using three examples drawn directly from the source dataset. These examples were selected to represent the main decision categories used during screening. The first example shows an acceptable numerical case, since questions containing numbers were often difficult to distinguish as acceptable or unacceptable. The second example shows a question containing one of the exclusion cases discussed above and illustrates a case that should be dropped. The third example shows a clear and concise question that can be kept.

Each example includes the question ID, question stem, answer options, correct answer, screening decision (yes, no, or maybe), and a short justification for the decision. This design exposes the model to the range of structural patterns it may encounter before classifying the full dataset. The full prompt is shown in Figure~\ref{fig:screening-prompt}.

\begin{figure}[!t]
\begin{tcolorbox}[
  colback=boxbg,
  colframe=headernavy,
  coltitle=white,
  fonttitle=\normalsize\bfseries,
  title=Screening Prompt,
  arc=4pt,
  boxrule=1pt,
  left=6pt, right=6pt,
  top=4pt, bottom=4pt,
  fontupper=\small,
]
\setlength{\parskip}{4pt}
\setlength{\parindent}{0pt}
You are a medical Arabic MCQA quality reviewer. Your task is to evaluate each question from the MedArabBench dataset and decide whether it should be KEPT or DROPPED for a medical answer generation benchmark.\\
 
\textbf{Your Job}\\
For each question, output one of three decisions:\\
\texttt{yes} \textrightarrow{} Keep it. Clear, well-formed question with a single unambiguous answer.\\
\texttt{no} \textrightarrow{} Drop it. Falls into one of the problematic categories below.\\
\texttt{maybe} \textrightarrow{} Borderline. Has a minor issue but could still be usable.\\
 
\textbf{Rules: When to DROP }\\
Drop a question if it matches ANY of the following:\\
\quad 1.\ Exclusion-logic phrasing
  (\textit{\textarabic{عدا}} / \textit{\textarabic{إلا}} / \textit{\textarabic{ما يلي}} / \textit{\textarabic{عدا ما}})\\
\quad 2.\ Options referencing each other
  (all of the above, A+B, \textit{\textarabic{كل ما سبق}})\\
\quad 3.\ Indirect/vague phrasing asking which statement is WRONG\\
\quad 4.\ Answer or options primarily in English\\
\quad 5.\ Question text primarily in English\\
\quad 6.\ Item merely states a label rather than posing a question\\

\textbf{Rules: When to FLAG as MAYBE}

Flag as \texttt{maybe} ONLY if:
\begin{enumerate}
    \item \textbf{Mixed language} — question is Arabic but one or two options contain English terms mixed in, not fully English.
    \item \textbf{Answer contains numbers with units in English.}
\end{enumerate}

\textbf{Rules: When to KEEP (yes)}

Keep if:
\begin{itemize}
    \item Question is in clear Arabic.
    \item All options are in Arabic, or are numbers, including integers, decimals, ratios, measurements, or medical abbreviations.
    \item There is exactly one correct answer.
    \item The question is direct, not asking ``which is WRONG'' or ``except''.
    \item No cross-referencing between options.
\end{itemize}

\textbf{Output Format}

Return ONLY a JSON array. Each element must have exactly two keys:
\begin{itemize}
    \item \texttt{id} — integer, from input.
    \item \texttt{keep} — \texttt{yes}, \texttt{no}, or \texttt{maybe}.
\end{itemize}

Do NOT include any explanation, preamble, or markdown. Output raw JSON only.
 
\textbf{Output Format}\\
Return ONLY a JSON array:
\texttt{[\{"id": <int>, "keep": "yes"|"no"|"maybe"\}, \ldots]}
\end{tcolorbox}
\caption{Screening prompt used to classify MCQA items for conversion to open-ended questions.}
\label{fig:screening-prompt}
\end{figure}

\subsubsection{Manual Verification}
\label{app:manual-verify}
 
We manually reviewed every label from the automated screening pass to verify its correctness and to adjudicate items flagged as borderline. Of the 4,958 source items, 4,014 were dropped as structurally incompatible with open-ended generation, leaving a retained set of 944 questions. This sharp reduction is deliberate: a large fraction of the source corpus consists of items whose structure is designed specifically for multiple-choice testing, and reformulating them as free-form questions would distort what the question is intended to test. Filtering aggressively at this stage allows the resulting benchmark to prioritize question quality over scale. Representative examples of dropped items, together with the justification for each exclusion, are shown in Table~\ref{tab:dropped}.
 
\begin{table*}[t]
\centering
\renewcommand{\arraystretch}{1.4}
\setlength{\tabcolsep}{9pt}
\begin{tabular}{p{3.3cm} p{3.3cm} p{5.4cm}}
\toprule
\textbf{Question} & \textbf{Answer} & \textbf{Justification for Exclusion} \\
\midrule
\parbox[t]{3.3cm}{
{\RaggedLeft\textarabic{العضلة الحرقفية :}\par}
\vspace{1.5pt}
\textit{\footnotesize The iliopsoas muscle:}
}
&
\parbox[t]{3.3cm}{
{\RaggedLeft\textarabic{تعمل على بسط الفخذ على البطن.}\par}
\vspace{1.5pt}
\textit{\footnotesize Acts to extend the thigh onto the abdomen.}
}
&
Not a well-formed question, merely states a body part without posing a query.
\\
\midrule
\parbox[t]{3.3cm}{
{\RaggedLeft\textarabic{الطبقة الثالثة من عضلات أخمص القدم تشمل عدا:}\par}
\vspace{1.5pt}
\textit{\footnotesize The third layer of plantar foot muscles includes, except:}
}
&
\parbox[t]{3.3cm}{
{\RaggedLeft\textarabic{باسطة الإصبع}\par}
\vspace{1.5pt}
\textit{\footnotesize Extensor digitorum brevis}
}
&
Exclusion-type question using \textarabic{عدا} (``except'') phrasing, tests identification of the one item that does not belong, a structure incompatible with open-ended answer generation.
\\
\midrule
\parbox[t]{3.3cm}{
{\RaggedLeft\textarabic{العبارة الخاطئة عن الشرايين المرنة هي :}\par}
\vspace{1.5pt}
\textit{\footnotesize The incorrect statement about elastic arteries is:}
}
&
\parbox[t]{3.3cm}{
{\RaggedLeft\textarabic{لونها مائل للأصفر بسبب تراكم الكولسترول.}\par}
\vspace{1.5pt}
\textit{\footnotesize Its colour is yellowish due to cholesterol accumulation.}
}
&
Asks the test-taker to identify the incorrect statement, inherently requires a closed option set and cannot be recast as a direct knowledge question.
\\
\midrule
\parbox[t]{3.3cm}{
{\RaggedLeft\textarabic{من وظائف الهرمونات التي يفرزها المبيض :}\par}
\vspace{1.5pt}
\textit{\footnotesize Among the functions of hormones secreted by the ovary:}
}
&
\parbox[t]{3.3cm}{
{\RaggedLeft\textarabic{كل ما سبق صحيح}\par}
\vspace{1.5pt}
\textit{\footnotesize All of the above is correct.}
}
&
Correct answer is ``all of the above'', a response meaningful only within a multiple-choice context and cannot serve as a standalone open-ended answer.
\\
\bottomrule
\end{tabular}
\caption{Representative MCQA items dropped during preprocessing, with the rationale for each exclusion. Each row illustrates a distinct failure mode that makes the item unsuitable for open-ended answer generation.}
\label{tab:dropped}
\end{table*}

\subsubsection{Automated Question Rewriting}
\label{app:reformulation}
 
We rewrote the 944 retained questions into clean, standalone open-ended form using \texttt{claude-opus-4-5-20250901} with one-shot prompting. Stems already phrased as direct interrogatives received only minor surface-level cleanup. Incomplete sentences, fill-in-the-blank prompts, and bare labels were recast as full natural Arabic questions. The reformulation prompt is shown in Figure~\ref{fig:reformulation-prompt-1}.
 
\begin{figure}[!t]
\begin{tcolorbox}[
  colback=boxbg,
  colframe=headernavy,
  coltitle=white,
  fonttitle=\normalsize\bfseries,
  title=Rewriting Prompt,
  arc=4pt,
  boxrule=1pt,
  left=6pt, right=6pt,
  top=4pt, bottom=4pt,
  fontupper=\small,
]
\setlength{\parskip}{4pt}
\setlength{\parindent}{0pt}
You are helping prepare an Arabic medical question answering benchmark.
You will be given an Arabic medical question stem originally designed as
a multiple choice question (MCQ). Your task is to rewrite it into a
clean, standalone open-ended question suitable for short answer
generation.\\
 
\textbf{Rules:}\par
\quad 1.\ Do NOT change the medical meaning, topic, or correct answer in any way.\par
\quad 2.\ Do NOT add information that is not in the original stem.\par
\quad 3.\ If the stem already starts with an Arabic interrogative
(\textit{\textarabic{ما}},
\textit{\textarabic{ماذا}},
\textit{\textarabic{كيف}},
\textit{\textarabic{من}},
\textit{\textarabic{أين}},
\textit{\textarabic{متى}},
\textit{\textarabic{كم}},
\textit{\textarabic{أي}}),
keep it as-is with only minimal cleanup.\par
\quad 4.\ If the stem is an incomplete sentence, label, or
fill-in-the-blank, rewrite it as a full natural Arabic question.\par
\quad 5.\ Remove any leading punctuation artifacts such as a lone period or colon.\par
\quad 6.\ Output ONLY the rewritten Arabic question --- no explanation, no preamble.\par

\medskip

\textbf{Example:}\par
\textbf{Input:}~~\textarabic{العصب الذي ينقل إحساس الذوق من ثلث اللسان الخلفي هو}\par
\textbf{Output:}~\textarabic{ما هو العصب الذي ينقل إحساس الذوق من ثلث اللسان الخلفي؟}

\end{tcolorbox}
\caption{One-shot prompt used to reformulate retained MCQA stems as standalone open-ended Arabic questions suitable for short-answer generation.}
\label{fig:reformulation-prompt-1}
\end{figure}
 
\subsubsection{Manual Question Review}
\label{app:quality-pass}

We then reviewed all 944 reformulated questions manually to verify correct formatting, clarity of phrasing, and fidelity to the original item's intent. Items that were ambiguous, unnaturally phrased, or inconsistent with the expected correct answer were flagged for correction. Of the 944 questions, 16 were revised and 4 were removed, yielding a final dataset of 940 open-ended questions used for the Short Answer Generation task.
 
A common pattern observed during this review was that the
LLM-generated reformulations, while grammatically well-formed, were
sometimes insufficiently specific. In these cases, the model produced
a broad open-ended question that, although technically valid, would
not constrain the respondent toward the particular aspect of knowledge
being tested. The annotated revision restored that specificity by
targeting the precise dimension of the answer the original item was
designed to elicit. Table~\ref{tab:rephrasing} illustrates this
pattern with a representative example.
 

\begin{table*}[!ht]
\centering
\small
\renewcommand{\arraystretch}{1.4}
\setlength{\tabcolsep}{10pt}
\begin{tabular}{p{3.2cm}
                p{3.5cm}
                p{5.0cm}}
\toprule
\multicolumn{1}{c}{\textbf{Original MCQA Stem}} & 
\multicolumn{1}{c}{\makecell{\textbf{LLM-Generated}\\\textbf{Rephrasing}}} & 
\multicolumn{1}{c}{\makecell{\textbf{Manually Annotated}\\\textbf{Rephrasing}}} \\
\midrule
\parbox[t]{3.2cm}{%
{\RaggedLeft\textarabic{حديبات مونتغومري هي غدد}\par}
\vspace{1.5pt}
\textit{\footnotesize Montgomery tubercles are glands}%
}
&
\parbox[t]{3.5cm}{%
{\RaggedLeft\textarabic{ما هي حديبات مونتغومري؟}\par}
\vspace{1.5pt}
\textit{\footnotesize What are Montgomery tubercles?}%
}
&
\parbox[t]{5.0cm}{%
{\RaggedLeft\textarabic{إلى أي صنف من أصناف الغدد تنتمي حديبات مونتغومري؟}\par}
\vspace{1.5pt}
\textit{\footnotesize To which type of glands do Montgomery tubercles belong?}%
}
\\
\bottomrule
\end{tabular}
\caption{Example of a question revised during the manual quality pass.
The LLM-generated phrasing admits a broad range of responses, whereas
the annotated version targets the specific medical fact tested by the
original item (correct answer: \textarabic{دهنية}, sebaceous).}
\label{tab:rephrasing}
\end{table*}

\subsection{Multi-Turn Dialogue Data Collection}
\label{app:task3-data-collection}

This section details the creation of the clinician-authored evaluation dataset used for the Multi-Turn Clinical Dialogue task. The dataset comprises 100 dialogues, each built from a structured clinical case template, and was developed in four stages: case template authoring, LLM-assisted dialogue generation, reference answer collection, and inter-annotator agreement assessment.

\subsubsection{Case Template Creation}
\label{app:case-templates}

We collaborated with three native Arabic-speaking physicians from a multi-specialty hopsital, each from a distinct specialty: critical care, pulmonary medicine, and general surgery. Clinicians were given an annotation guide specifying the task: to author clinical case templates that would serve as the basis for generating multi-turn patient–assistant dialogues in Modern Standard Arabic (MSA). 

Case distribution was determined in consultation with the clinicians to reflect the range of scenarios commonly encountered in clinical settings. Templates were required to span nine organ systems and to cover a range of scenario types, including diagnosis clarification, emergency referral, medication guidance, and preventive counseling. Each template followed a fixed schema with six fields, as shown in the representative example in Table~\ref{tab:template-example-1}. The resulting distribution of cases across organ systems is summarised in Table~\ref{tab:organ-systems}.

\begin{table}[!ht]
\centering
\small
\renewcommand{\arraystretch}{1.4}
\setlength{\tabcolsep}{12pt}
\begin{tabular}{lc}
\toprule
\textbf{Organ System} & \textbf{Number of Cases} \\
\midrule
Abdomen          & 15 \\
Cardiovascular   & 15 \\
Endocrinology    & 15 \\
Lymphatic System & 5  \\
Musculoskeletal  & 15 \\
Nervous System   & 10 \\
Respiratory      & 15 \\
Urinary/Kidney   & 5  \\
Urology          & 5  \\
\midrule
\textbf{Total}   & \textbf{100} \\
\bottomrule
\end{tabular}
\caption{Distribution of the 100 clinician-authored cases across nine organ systems. Coverage was weighted toward systems most commonly encountered in clinical dialogue.}
\label{tab:organ-systems}
\end{table}

\begin{table*}[!htbp]
\centering
\small
\renewcommand{\arraystretch}{1.4}
\setlength{\tabcolsep}{10pt}
\begin{tabular}{>{\raggedright\arraybackslash}p{3cm}
                >{\raggedright\arraybackslash}p{10cm}}
\toprule
\textbf{Field} & \textbf{Content} \\
\midrule
Organ System & Endocrinology \\
\midrule
Problem Type & Clinic Non-Emergency \\
\midrule
Patient Profile &
A 64-year-old postmenopausal female with a history of osteopenia.
She has no previous history of neck surgery or radiation exposure. \\
\midrule
Symptoms \& History &
The patient reports vague aching in her bones and recurrent bouts of
constipation over several months. She has experienced two episodes of
painful kidney stones in the last three years. Physical exam is
unremarkable, but she mentions feeling increasingly foggy and fatigued.
Laboratory tests reveal a persistently elevated serum calcium of
10.8~mg/dL. Her vitamin D levels are within the normal range. \\
\midrule
Red Flag Symptoms & Chronic hypercalcemia. \\
\midrule
Primary Reasoning Objective &
Diagnosis of primary hyperparathyroidism, elevated calcium with
inappropriately normal or high PTH suggests a parathyroid adenoma. \\
\bottomrule
\end{tabular}
\caption{Representative example of a completed clinical case template (Case~1, Endocrinology, Clinic Non-Emergency), authored by one of the three clinicians.}
\label{tab:template-example-1}
\end{table*}

\subsubsection{LLM-Assisted Dialogue Generation}
\label{app:dialogue-gen}

We prompted GPT-5.2 to generate a multi-turn dialogue for each of the 100 clinician-authored templates. Each dialogue was a patient–assistant exchange in MSA, constrained to span three to eight turns and to conclude with a patient question. This structure follows HealthBench~\citep{arora2025healthbenchevaluatinglargelanguage}: the dialogue terminates with a user turn so that, at evaluation time, the model under test must produce the final assistant response. 

\begin{figure*}[!htbp]
    \centering
    \begin{adjustbox}{width=\textwidth, center}
        \includegraphics{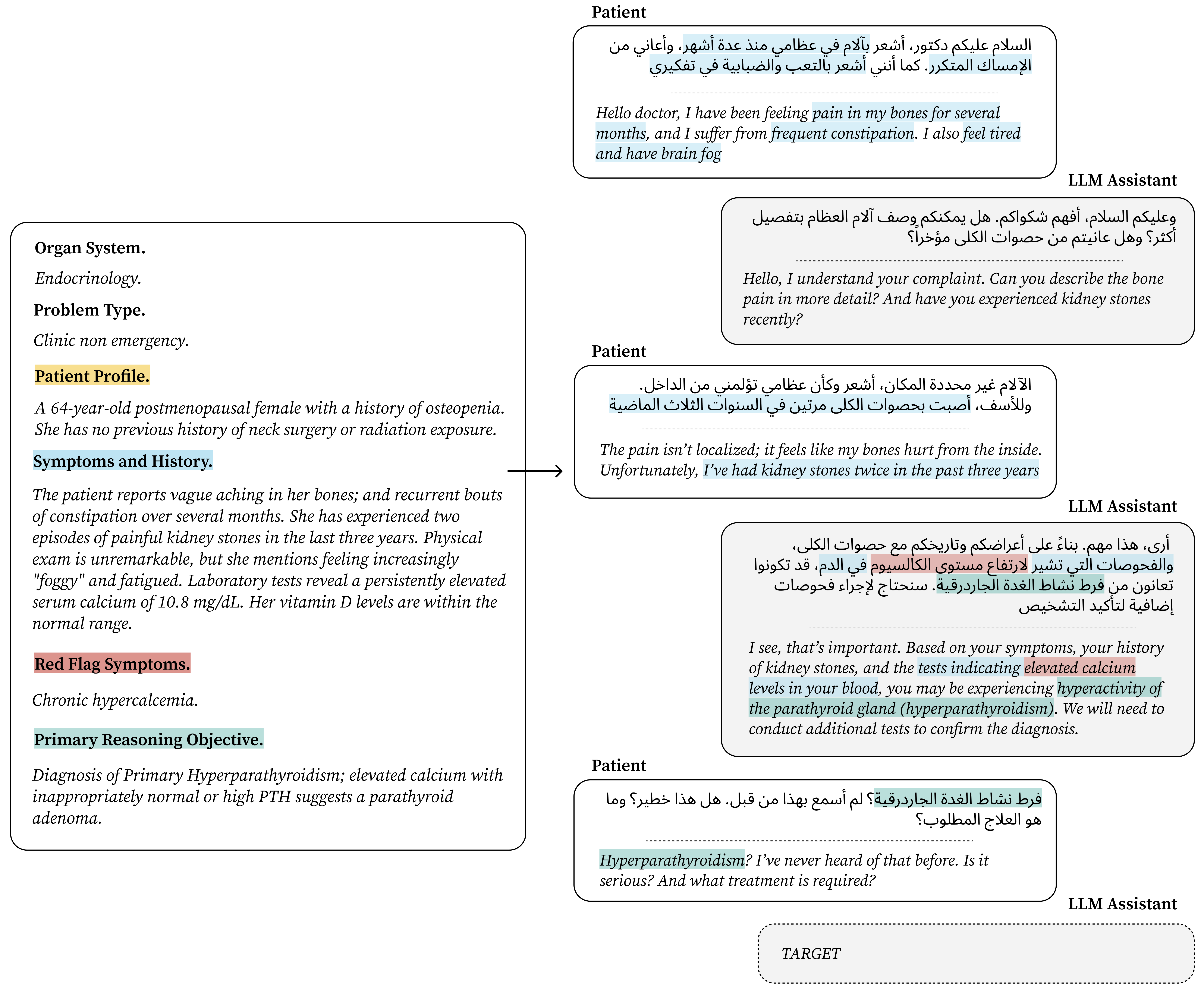}
    \end{adjustbox}
    \caption{Example from our multi-turn clinical dialogue dataset. Each dialogue is grounded in a structured clinical scenario authored by physicians (left). Colored highlights mark spans where \textbf{symptoms}, \textbf{red-flag indicators}, and \textbf{reasoning objectives} from the scenario appear in the turns. Given turns $T_1, \ldots, T_{n-1}$, the model must generate $T_n$, evaluated in both Arabic and English. \textbf{Patient-profile considerations} may not surface lexically in the dialogue but must be reflected in the generated turn.}
    \label{fig:dialogue-example-1}
\end{figure*}

The model was instructed to seek clarification naturally when needed, to avoid premature clinical escalation, and to refrain from introducing numerical thresholds or clinical data not grounded in the template. The final assistant turn, the diagnostic response, was deliberately withheld at this stage and provided independently by the clinicians. The dialogue generation prompt is shown in Figure~\ref{fig:dialogue-gen-prompt}, and Figure~\ref{fig:dialogue-example-1} shows a completed template with its generated dialogue.

\begin{figure}[!t]
\begin{tcolorbox}[
  colback=boxbg,
  colframe=headernavy,
  coltitle=white,
  fonttitle=\normalsize\bfseries,
  title=Dialect Translation Prompt,
  arc=4pt,
  boxrule=1pt,
  left=6pt, right=6pt,
  top=4pt, bottom=4pt,
  fontupper=\small,
]
\setlength{\parskip}{4pt}
\setlength{\parindent}{0pt}
You are a professional medical translator and a native speaker of each
target Arabic dialect. Translate the following Modern Standard Arabic
(MSA) medical conversation into four spoken Arabic dialects. The output
must sound like how a real speaker of that dialect would actually talk
to their doctor --- not MSA with a few dialect particles sprinkled in.\\
 
\textbf{Translate into:} (1) Emirati \quad (2) Jordanian \quad
(3) Moroccan Darija \quad (4) Egyptian\\
 
\textbf{Core rules (non-negotiable):}\\
\quad 1.\ Preserve all medical meaning exactly. Do NOT add, remove, or
  simplify medical content.\\
\quad 2.\ Keep the same speaker turns, urgency, and triage tone.\\
\quad 3.\ Do not mix dialects within a single translation.\\
\quad 4.\ Medical terminology: use English or French if more common in
  that dialect.\\
 
\textbf{Authenticity rules:}\par
\quad --\ Default to spoken register, not written register.\par
\quad --\ Replace MSA connectors and fillers with dialectal equivalents.\par
\quad --\ Replace abstract MSA phrasing with concrete dialectal phrasing\par
\quad\quad (e.g., \textit{\textarabic{ضبابية في التفكير}}
  \(\rightarrow\) Darija: \textit{\textarabic{ما كنركزش مزيان}};
  Egyptian: \textit{\textarabic{مخي مش راكز}};\par
\quad\quad Jordanian: \textit{\textarabic{بحس حالي مش مركز}};
  Emirati: \textit{\textarabic{ما أقدر أركّز}}).
 
\textbf{Self-check before outputting:} Re-read your translation and
ask: ``Would a real patient in a clinic actually say this sentence to
their doctor, or does it sound like an MSA sentence wearing a dialectal
costume?'' If the latter, rewrite it.\\
 
\textbf{Output format:}\\
Emirati Arabic: [Translated conversation]\\
Jordanian Arabic: [Translated conversation]\\
Moroccan Darija: [Translated conversation]\\
Egyptian Arabic: [Translated conversation]\\
 
\textbf{Text to translate:} [INSERT MSA CONVERSATION HERE]
\end{tcolorbox}
\caption{Prompt used to translate MSA dialogues into four regional
Arabic dialects. }
\label{fig:dialect-prompt}
\end{figure}

\begin{figure}[!t]
\begin{tcolorbox}[
  colback=boxbg,
  colframe=headernavy,
  coltitle=white,
  fonttitle=\normalsize\bfseries,
  title=Dialogue Generation Prompt,
  arc=4pt,
  boxrule=1pt,
  left=6pt, right=6pt,
  top=4pt, bottom=4pt,
  fontupper=\small,
]
\setlength{\parskip}{4pt}
\setlength{\parindent}{0pt}
You are generating a realistic medical conversation in Arabic between a
patient and a medical assistant chatbot.\\
 
Use ONLY the information provided in the case template below. Do not
introduce new medical history, vital signs, or test results unless they
logically emerge through dialogue clarification.\\
 
\textbf{Requirements:}\\
\quad 1.\ Generate a natural, realistic conversation in Modern Standard Arabic.\\
\quad 2.\ The conversation must be between a patient and a medical assistant.\\
\quad 3.\ Length must be between 3 and 8 total turns.\\
\quad 4.\ The last turn MUST be a patient question.\\
\quad 5.\ The assistant should appropriately seek clarification when necessary.\\
\quad 6.\ Do not immediately escalate unless clearly justified by the evolving dialogue.\\
\quad 7.\ Do not introduce numeric thresholds or new clinical data not grounded in the template.\\
\quad 8.\ The conversation should feel authentic and medically plausible.\\
 
\textbf{Output format:}\\
Patient: \ldots\\
Assistant: \ldots\\
Patient: \ldots\\
\ldots~(end with Patient message)\\
 
\textbf{Case Template:} [INSERT TEMPLATE HERE]
\end{tcolorbox}
\caption{Prompt used to generate multi-turn patient--assistant dialogues
from the clinician-authored case templates.}
\label{fig:dialogue-gen-prompt}
\end{figure}

\subsubsection{Reference Answer Collection}
\label{app:reference-answers}
 
To obtain ground-truth final-turn responses for each dialogue, the three clinicians were asked to independently review each case and submit the best possible response. To ensure that each case received two independent reference answers, the 100 dialogues were distributed such that each case was reviewed by exactly two different clinicians. Responses were collected via a structured Google Form presenting each clinician with the case identifier, organ system, problem type, primary reasoning objective, and the full dialogue.

The clinicians elected to record their responses as audio rather than typed text. The resulting recordings were transcribed using Whisper-Large-V3, and each transcription was then verified by reading the text against its source recording, with corrections applied where necessary. Table~\ref{tab:reference-answers} shows a representative reference-answer pair.
 
\begin{table*}[!htbp]
\centering
\small
\renewcommand{\arraystretch}{1.4}
\setlength{\tabcolsep}{8pt}
\begin{tabular}{>{\centering\arraybackslash}p{1.2cm}
                p{5.3cm}
                p{5.3cm}}
\toprule
\textbf{Case} & \multicolumn{1}{c}{\textbf{Response A}} &
\multicolumn{1}{c}{\textbf{Response B}} \\
\midrule
1 &
\parbox[t]{5.3cm}{%
{\RaggedLeft\textarabic{الغدة الجار دراقية صغيرة تقع خلف الغدة الدراقية. الوظيفة الأساسية هي تحديد الكالسيوم في الدم. ممكن أن تكون خطيرة إذا الكالسيوم ارتفع بشكل حاد. الخبر الجيد أن هذا المرض يمكن أن يعالج عن طريق إقصاء هذه الغدد بطريقة جراحية. من دون علاج ممكن أن تؤدي إلى أوجاع حصى كلسية وبطء في حركة الجهاز الهضمي.}\par}
\vspace{1.5pt}
\textit{\footnotesize The parathyroid gland is small and sits behind the thyroid. Its main role is regulating blood calcium. It can be dangerous if calcium rises sharply. Treatment is typically surgical. Untreated it causes kidney stones and slowed digestion.}%
}
&
\parbox[t]{5.3cm}{%
{\RaggedLeft\textarabic{فرط نشاط جارات الدرقية ليس مرضاً مميتاً بل يمكن علاجه. أولاً يجب تحديد نوع فرط نشاط جارات الدرقية وسبب المرض. عادةً ما يكون العلاج جراحياً.}\par}
\vspace{1.5pt}
\textit{\footnotesize Hyperparathyroidism is not fatal and is treatable. The type and underlying cause must first be identified. Treatment is typically surgical.}%
}
\\
\bottomrule
\end{tabular}
\caption{Representative example of two independently collected reference
answers for Case~1 (Endocrinology, Clinic Non-Emergency:
hyperparathyroidism). English translations are shown in italics beneath each Arabic response.}
\label{tab:reference-answers}
\end{table*}

\subsubsection{Inter-Annotator Agreement}
\label{app:iaa}

Before writing each response, clinicians were provided with the case template and dialogue, and instructed to produce the final assistant turn according to the following rubric:

\begin{itemize}
    \item Be clinically accurate.
    \item Be written in Modern Standard Arabic (MSA).
    \item Be 4--6 sentences in length.
    \item Focus on the primary clinical reasoning objective of the case.
    \item Include both a diagnosis, or explanation when appropriate, and an appropriate triage recommendation.
\end{itemize}
 
To assess agreement between the two reference answers collected per case, a fourth Arabic-speaking clinician from a general hospital, reviewed all 100 dialogue–response pairs, labeling each as Agreement when both responses reached the same diagnosis, and Disagreement otherwise. For disagreements, the reviewer also indicated the preferred response. Table~\ref{tab:iaa-example} shows representative examples. Of the 100 reviewed cases, 87 were labeled as agreement; for the remaining 13 disagreement cases, the reviewer indicated a preferred response or authored a revised gold-standard answer when neither response was satisfactory.

To assess agreement between the two reference answers collected per case, a fourth Arabic-speaking clinician from a general hospital reviewed all 100 dialogue--response pairs, labeling each pair as Agreement when both responses reached the same diagnosis, and Disagreement otherwise. For disagreements, the reviewer also indicated the preferred response. Table~\ref{tab:iaa-example} shows representative examples. Of the 100 reviewed cases, 87 were labeled as agreement; for the remaining 13 disagreement cases, the reviewer indicated a preferred response or authored a revised gold-standard answer when neither response was satisfactory.

\begin{table*}[t]
\centering
\small
\renewcommand{\arraystretch}{1.3}
\setlength{\tabcolsep}{5pt}
\begin{tabular}{>{\centering\arraybackslash}p{0.7cm}
                p{4.0cm}
                p{4.0cm}
                p{3.0cm}}
\toprule
\textbf{Case} &
\multicolumn{1}{c}{\textbf{Response A}} &
\multicolumn{1}{c}{\textbf{Response B}} &
\multicolumn{1}{c}{\textbf{Agreement}} \\
\midrule
1 &
\parbox[t]{4.0cm}{%
{\RaggedLeft\textarabic{الغدة الجار دراقية صغيرة تقع خلف الغدة الدراقية. الوظيفة الأساسية هي تحديد الكالسيوم في الدم، ممكن أن تكون خطيرة إذا الكالسيوم ارتفع بشكل حاد. الخبر الجيد أن هذا المرض يمكن أن يعالج عن طريق إقصاء هذه الغدد بطريقة جراحية، المهم أن لا تنتظر وعليك أن تتابع مع متخصص للغدد الجار دراقية. من دون علاج ممكن أن تؤدي إلى أوجاع حصى كلسية وبطء في حركة الجهاز الهضمي.}\par}
\vspace{1.5pt}
\textit{\footnotesize The parathyroid gland is small and sits behind the thyroid. Its main function is regulating blood calcium, which can be dangerous if it rises sharply. Treatment is surgical; without treatment it causes kidney stones and slowed digestion.}%
}
&
\parbox[t]{4.0cm}{%
{\RaggedLeft\textarabic{فرط نشاط جارات الدرقية ليس مرضاً مميتاً، بل يمكن علاجه. أولاً، يجب تحديد نوع فرط نشاط جارات الدرقية وسبب المرض. عادةً ما يكون العلاج جراحياً.}\par}
\vspace{1.5pt}
\textit{\footnotesize Hyperparathyroidism is not fatal and is treatable. The type and underlying cause must first be identified. Treatment is typically surgical.}%
}
&
Agreement \\
\midrule
8 &
\parbox[t]{4.0cm}{%
{\RaggedLeft\textarabic{أنصحك بأن تتابع مع دكتور متخصص في الغدد لتأكيد المرض. ممكن تحتاج هورمون النمو وتأخذ أسابيع لعدة أشهر لتعود مثل ما كنت من قبل، في الشعور التحسن في العضلات وانخفاض في الدهون وممارسة الحياة الاجتماعية العادية، ولكنه مهم أن تتابع مع دكتور متخصص بهذه الأمراض.}\par}
\vspace{1.5pt}
\textit{\footnotesize I recommend following up with an endocrinologist to confirm the diagnosis. You may need growth hormone, and it can take weeks to months before improvements in muscle mass, reduced fat, and normal social functioning are felt.}%
}
&
\parbox[t]{4.0cm}{%
{\RaggedLeft\textarabic{بعد تأكيد التشخيص من خلال فحص الدم، الذي يستغرق عادةً من ساعتين إلى ثلاث ساعات، حيث نراقب استجابة جسمك للأنسولين وانخفاض سكر الدم، وما إذا كان يستجيب بإفراز الكورتيزول. قد تشعر بأعراض نقص سكر الدم، كالدوار والتعرق والإرهاق. إذا تأكد التشخيص، فستحتاج إلى حقن هرمون النمو لمدة تتراوح بين ستة وتسعة أشهر على الأقل}\par}
\vspace{1.5pt}
\textit{\footnotesize After confirming via blood test (2--3 hours), monitoring insulin response and cortisol. Side effects include dizziness, sweating, and fatigue. Growth hormone injections required for at least 6--9 months.}%
}
&
Disagreement, Response~B preferred \\
\bottomrule
\end{tabular}
\caption{Examples of inter-annotator agreement annotation. Case~1 (hyperparathyroidism) illustrates an agreement
outcome, while case~8 illustrates a disagreement outcome. English translations are shown in italics beneath each Arabic response.}
\label{tab:iaa-example}
\end{table*}
 
\subsubsection{Dialect Translation}
\label{app:dialect-translation}
 
To extend the utility of the dataset beyond Modern Standard Arabic and
support evaluation across dialectally diverse user populations, the
100 MSA dialogues were translated into four regionally representative
Arabic dialects: Emirati (Gulf), Jordanian (Levant), Moroccan Darija
(Maghreb), and Egyptian. 

\begin{figure*}[!htbp]
\centering
\includegraphics[width=\textwidth]{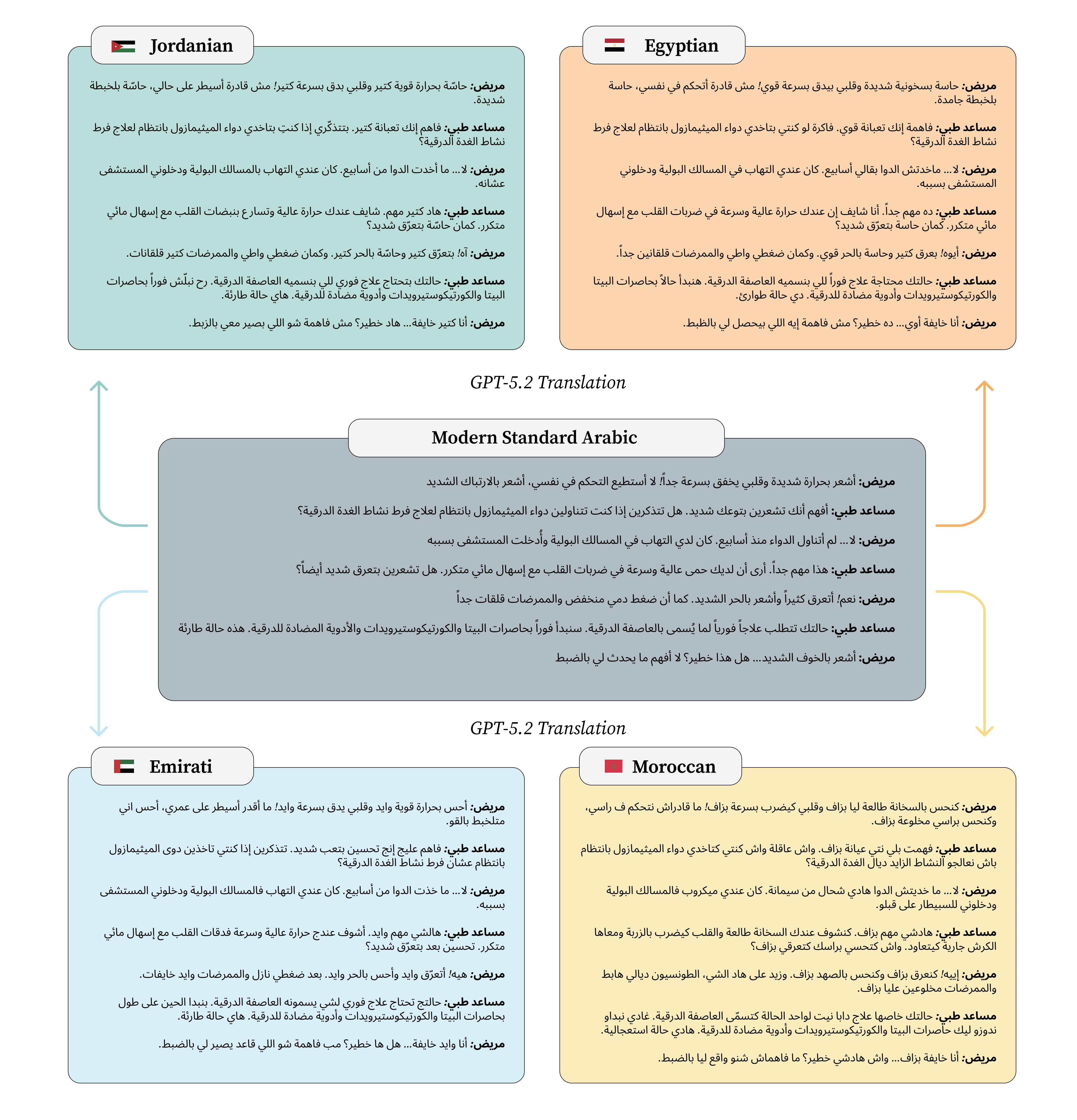}
\caption{Sample dialogue in MSA (center) and its four dialect translations obtained with GPT-5.2: Jordanian and Egyptian (top), Emirati and Moroccan Darija (bottom).}
\label{fig:dialect-example}
\end{figure*}

Translations were generated using GPT-5.2, prompted to produce authentic dialectal speech that would be natural in a real clinical interaction. The prompt encouraged the use of authentic dialectal vocabulary and phrasing, including code-switching to English or French medical terminology where appropriate for a given dialect. The translation prompt is reproduced in Figure~\ref{fig:dialect-prompt}, and Figure~\ref{fig:dialect-example} shows a sample dialogue in MSA and its four dialect translations.

For each dialect, the 100 samples were divided evenly between two native speakers who reviewed the generated translations and corrected expressions that were unnatural or clinically inaccurate. The extent of editing varied substantially across dialects: Moroccan Darija required the most extensive corrections, as the model often produced MSA-dominant text with only sporadic dialect-specific lexical insertions rather than authentic Darija. Table~\ref{tab:dialect-example} shows representative LLM-generated translations and their human-corrected versions.
 
\begin{table*}[!htbp]
\centering
\small
\renewcommand{\arraystretch}{1.4}
\setlength{\tabcolsep}{7pt}
\begin{tabular}{>{\raggedright\arraybackslash}p{1.5cm}
                p{5.3cm}
                p{5.3cm}}
\toprule
\textbf{Dialect} &
\multicolumn{1}{c}{\textbf{LLM-Generated Translation}} &
\multicolumn{1}{c}{\textbf{Human-Corrected Translation}} \\
\midrule
Emirati &
\parbox[t]{5.3cm}{%
{\RaggedLeft\textarabic{مريض: السلام عليكم دكتور، من كم شهر وأنا أحس بآلام في عظامي، وعندي إمساك بشكل متكرر. بعد أحس بتعب وما أقدر أركّز، تفكيري مو صافي.}\par}
\vspace{1.5pt}
\textit{\footnotesize Patient: I've had pain in my bones for months and keep getting constipated. I also feel tired and can't focus, my thinking isn't clear.}%
}
&
\parbox[t]{5.3cm}{%
{\RaggedLeft\textarabic{مريض: السلام عليكم دكتور، من كم شهر وأنا أحس بالعوار في عظامي، وعندي إمساك بشكل متكرر. بعد أحس بتعب وما أقدر أركّز، تفكيري مب صافي.}\par}
\vspace{1.5pt}
\textit{\footnotesize \textarabic{العوار} replaces \textarabic{آلام} for pain; \textarabic{مب} replaces \textarabic{مو} as negation.}%
}
\\
\midrule
Egyptian &
\parbox[t]{5.3cm}{%
{\RaggedLeft\textarabic{مريض: حاسة بألم شديد في بطني وغثيان مستمر، ومش قادرة أوقف ترجيع من بدري قوي من الصبح.}\par}
\vspace{1.5pt}
\textit{\footnotesize Patient: I have severe stomach pain and constant nausea, and I haven't been able to stop vomiting since early this morning.}%
}
&
\parbox[t]{5.3cm}{%
{\RaggedLeft\textarabic{مريض: حاسة بألم شديد في بطني وعلى طول حاسة إني عايزة أَرَجّع، ومش عارفة أبطّل ترجيع من الصبح.}\par}
\vspace{1.5pt}
\textit{\footnotesize \textarabic{على طول حاسة إني عايزة أَرَجّع} replaces \textarabic{غثيان مستمر} for nausea; \textarabic{مش عارفة أبطّل} replaces \textarabic{مش قادرة أوقف} for inability to stop.}%
}
\\
\midrule
Jordanian &
\parbox[t]{5.3cm}{%
{\RaggedLeft\textarabic{مريض: يعني الوضع خطير كتير؟ كمان حاسة بدوخة، كأني ممكن أغيب عن الوعي.}\par}
\vspace{1.5pt}
\textit{\footnotesize Patient: So is the situation very serious? I also feel dizzy, like I might lose consciousness.}%
}
&
\parbox[t]{5.3cm}{%
{\RaggedLeft\textarabic{مريض: يعني الوضع خطير كتير؟ كمان حاسة بدوخة، كأني ممكن يغمى علي.}\par}
\vspace{1.5pt}
\textit{\footnotesize \textarabic{يغمى علي} replaces \textarabic{أغيب عن الوعي} as the Jordanian-idiomatic expression for losing consciousness.}%
}
\\
\midrule
Moroccan &
\parbox[t]{5.3cm}{%
{\RaggedLeft\textarabic{مريضة: إييه، كنحس بالبرد ديما حتى إلا كان الجو سخون، وبشرتي ولات ناشفة بزاف. وزدت لاحظت الشعرة بدات كتطيح ليا من الجوانب.}\par}
\vspace{1.5pt}
\textit{\footnotesize Patient: Yes, I always feel cold even when the weather is warm, and my skin has become very dry. I also noticed my hair started falling out on the sides.}%
}
&
\parbox[t]{5.3cm}{%
{\RaggedLeft\textarabic{مريضة: إييه، كنحس بالبرد ديما حتى إلا كان الجو سخون، ولبشرة ديالي ولات ناشفة بزاف. وزدت لاحظت شعري بدا كيطيح ليا من الجنب.}\par}
\vspace{1.5pt}
\textit{\footnotesize \textarabic{لبشرة ديالي} replaces \textarabic{بشرتي} using Darija possessive construction; \textarabic{شعري} replaces \textarabic{الشعرة}; \textarabic{من الجنب} replaces \textarabic{من الجوانب}.}%
}
\\
\bottomrule
\end{tabular}
\caption{Representative examples of LLM-generated versus
human-corrected dialect translations across all four target dialects.
Each row illustrates a distinct dialectal correction pattern. English translations are
shown in italics beneath each Arabic example.}
\label{tab:dialect-example}
\end{table*}











\FloatBarrier 
\section{LLM-as-a-Judge Reliability} 
\label{app:llm-judge-reliability}

BERTScore-F1 is an imperfect proxy for clinical quality in open-ended generation: prior work has shown it may not reliably distinguish model quality in medical tasks~\citep{bedi2026medhelm}, and does not directly evaluate clinical correctness. We therefore adopt an LLM-as-a-judge approach as an additional metric for the Short Answer Generation task. To validate this choice, we collect human labels on a subset of Short Answer Generation outputs and measure how well both BERTScore-F1 and the LLM judge correlate with human assessment.

\paragraph{Validation Subset.} To balance annotation costs with statistical coverage, we construct the validation subset by randomly sampling 100 examples from the 940-example dataset with a fixed seed (42) to ensure reproducibility. For each selected example, we collect the predictions from 18 of the 23 evaluated models, excluding the five adaptation methods, as the goal of this validation is to assess agreement patterns across models rather than benchmark every evaluated model. This yields 1,800 model outputs for human validation.

\paragraph{Annotation Protocol.} To reduce annotator workload, we apply an exact-match pre-filter: model outputs that exactly match the reference answer are automatically labeled as correct, as their correctness is unambiguous. This accounts for 54 out of the 1800 total predictions in the validation subset. Each remaining (question, reference answer, model output) triple is independently evaluated by two medical students in their last year (6th year), who assign a binary label (correct or incorrect) following the rubric in Table~\ref{tab:human-annotation-rubric}. The LLM-judge labels are not shared with the annotators. 

Model-level human-judge accuracy (percentage correct) is computed as the percentage of predictions labeled correct, calculated separately for each annotator and then averaged. We note that approximately 6\% of the questions in the subset were flagged by both annotators as having incorrect ground truth, highlighting a limitation of the dataset.

\begin{table}[ht]
\centering
\small
\begin{tabular}{l p{0.72\linewidth}}
\specialrule{1pt}{0pt}{1.5pt}
\textbf{Label} & \textbf{Criteria} \\
\specialrule{1pt}{0pt}{1.5pt}
Correct & Medically accurate, answers the question, and conveys the same clinical meaning as the reference answer, regardless of wording. Additional information that does not contradict the reference is permitted. \\
\specialrule{0.4pt}{2pt}{2pt}
Incorrect & Contains factual medical errors, omits the essential answer, contradicts the reference, or fails to answer the question. \\
\specialrule{1pt}{0pt}{0pt}
\end{tabular}
\caption{Annotation rubric used by human annotators to label model outputs as correct or incorrect.}
\label{tab:human-annotation-rubric}
\end{table}

\begin{table*}[!htbp]
\centering
\small
\setlength{\tabcolsep}{5pt}
\renewcommand{\arraystretch}{1.15}
\resizebox{\textwidth}{!}{%
\begin{tabular}{>{\raggedright\arraybackslash}m{3.0cm}
                >{\raggedright\arraybackslash}m{4.7cm}
                c c c}
\specialrule{1pt}{0pt}{1.5pt}
\multicolumn{1}{c}{\textbf{Category}} & \textbf{Model} & \textbf{BERTScore-F1} & \textbf{LLM-Judge Accuracy} & \textbf{Human-Judge Accuracy} \\
\specialrule{1pt}{0pt}{1.5pt}

\multirow{3}{3.0cm}{\parbox[c]{3.0cm}{\centering Closed-Source\\General-Purpose LLMs}}
& GPT-5.2 & 63.0 & 62.0 & 84.0 \\
& Gemini-2.5-Flash & \textbf{64.5} & 62.0 & 86.0 \\
& Claude-Opus-4.6 & 60.1 & \textbf{64.0} & \textbf{87.5} \\

\specialrule{0.6pt}{0pt}{0pt}

\multirow{6}{3.0cm}{\parbox[c]{3.0cm}{\centering Open-Source\\General-Purpose\\LLMs}}
& Mistral-7B-Instruct-v0.3 & 52.2 & 2.0 & 6.5 \\
& Llama-3.1-8B-Instruct & 55.8 & 7.0 & 14.5 \\
& Mistral-Small-3.2-24B & \textbf{61.2} & 26.0 & 50.0 \\
& Gemma-3-27B-it & 58.9 & 31.0 & 52.5 \\
& Llama-3.3-70B-Instruct & 57.7 & 16.0 & 28.0 \\
& DeepSeek-V3.2 & 61.9 & \textbf{52.0} & \textbf{73.0} \\

\specialrule{0.6pt}{0pt}{0pt}

\multirow{6}{3.0cm}{\parbox[c]{3.0cm}{\centering Arabic/Multilingual\\General-Purpose\\LLMs}}
& Jais-2-8B-Chat & 54.2 & \textbf{23.0} & \textbf{45.5} \\
& ALLaM-7B-Instruct-preview & 56.4 & \textbf{23.0} & 39.0 \\
& Aya-Expanse-8B & 54.5 & 18.0 & 39.0 \\
& SILMA-9B-Instruct-v1.0 & \textbf{57.2} & 22.0 & 32.0 \\
& Falcon-H1-7B-Instruct & 56.6 & 19.0 & 33.0 \\
& Fanar-1-9B-Instruct & 51.6 & 9.0 & 12.0 \\

\specialrule{0.6pt}{0pt}{0pt}

\multirow{3}{3.0cm}{\parbox[c]{3.0cm}{\centering Medical LLMs}}
& MedGemma-27B-Text-It & \textbf{60.1} & \textbf{35.0} & \textbf{58.5} \\
& Meditron-3-70B & 53.3 & 27.0 & 51.0 \\
& Llama3-Med42-70B & 53.6 & 25.0 & 40.0 \\

\specialrule{1pt}{0pt}{0pt}
\end{tabular}%
}
\caption{Model-level evaluation scores on the 100-example validation subset.}
\label{tab:judge-validation-model-scores}
\end{table*}

\begin{figure*}[!htbp]
\centering
\includegraphics[width=\textwidth]{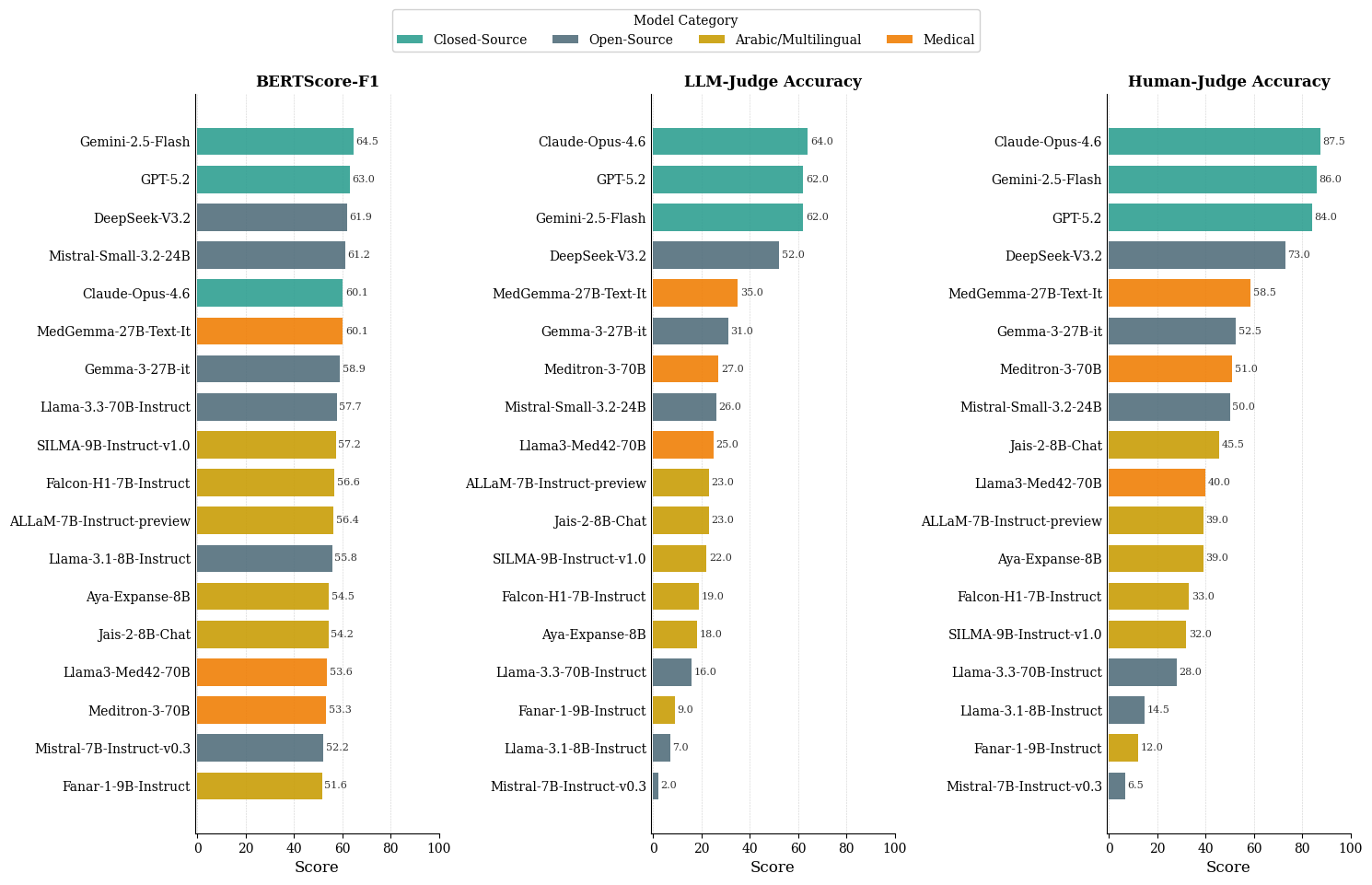}
\caption{Comparison of model rankings across BERTScore-F1, LLM-judge accuracy, and human-judge accuracy.}
\label{fig:validation-rank}
\end{figure*}

\paragraph{Results.} Table ~\ref{tab:judge-validation-model-scores} summarizes the BERTScore-F1, LLM-judge accuracy (\% of correct predictions), and human-judge accuracy (\% of correct predictions) on the validation subset of the MedArabench-OE dataset. Each of the three metrics produces an independent model ranking based on these aggregated scores, as shown in Figure~\ref{fig:validation-rank}. To assess agreement of LLM as a judge to human judgement, we compute both Pearson correlation and Spearman’s rank correlation across models. For comparison, we also compute the same correlations between BERTScore-F1 and human evaluation. We adopt a threshold of $\rho\ge0.8$ as a criterion for validating the judge. As shown in Table~\ref{tab:judge-correlation}, the LLM judge achieves substantially higher agreement with human judgments (Pearson: 0.978; Spearman: 0.982) than BERTScore-F1 (Pearson: 0.802; Spearman: 0.715), satisfying our criterion of $\rho\ge0.8$. 

\begin{table}[ht]
\centering
\small
\setlength{\tabcolsep}{8pt}
\renewcommand{\arraystretch}{1.15}
\resizebox{\columnwidth}{!}{%
\begin{tabular}{l c c}
\specialrule{1pt}{0pt}{1.5pt}
\textbf{Metric} & \textbf{Pearson} & \textbf{Spearman ($\rho$)} \\
\specialrule{1pt}{0pt}{1.5pt}

BERTScore-F1 & 0.802 & 0.715 \\
LLM-Judge Accuracy & \textbf{0.978} & \textbf{0.982} \\

\specialrule{1pt}{0pt}{0pt}
\end{tabular}%
}
\caption{Correlation with Human-Judge Accuracy on the 100-example validation subset.}
\label{tab:judge-correlation}
\end{table}



\paragraph{Bias Check: Self-Preference of the Judge.} The validation study further supports our inclusion of ChatGPT 5.2 as a generator model for the Short Answer Generation task. Its rank is similar under human-judge accuracy and LLM-judge accuracy (Figure~\ref{fig:validation-rank}), indicating no meaningful self-preference in the LLM-as-a-Judge evaluation.

\section{Training Methodology}
\label{app:lora_lr_search}

\paragraph{Fixed hyperparameters.}
Table~\ref{tab:fixed-hparams} lists all hyperparameters held constant
across every LoRA experiment reported in this paper.
The LoRA configuration ($r{=}16$, $\alpha_{\mathrm{LoRA}}{=}32$,
dropout~$0.05$) follows standard practice for instruction-tuned 24\,B-parameter
models~\citep{hu2022lora}.
 
\begin{table}[ht]
\centering
\setlength{\tabcolsep}{8pt}
\renewcommand{\arraystretch}{1.2}
\begin{tabular}{lc}
\toprule
\textbf{Hyperparameter} & \textbf{Value} \\
\midrule
LoRA rank ($r$)                   & 16 \\
LoRA $\alpha_{\mathrm{LoRA}}$     & 32 \\
LoRA dropout                      & 0.05 \\
Max sequence length               & 1{,}024 tokens \\
Training epochs                   & 10 \\
Early-stopping patience           & 1 checkpoint \\
Warmup ratio                      & 0.05 \\
LR scheduler                      & Cosine \\
Optimizer                         & AdamW \\
Precision                         & bfloat16 \\
GPUs                              & 2 $\times$ A100 80\,GB \\
\bottomrule
\end{tabular}
\caption{Hyperparameters held constant across all LoRA experiments.}
\label{tab:fixed-hparams}
\end{table}

\paragraph{Convergence analysis.}

\begin{figure}[ht]
\centering
\includegraphics[width=0.62\linewidth]{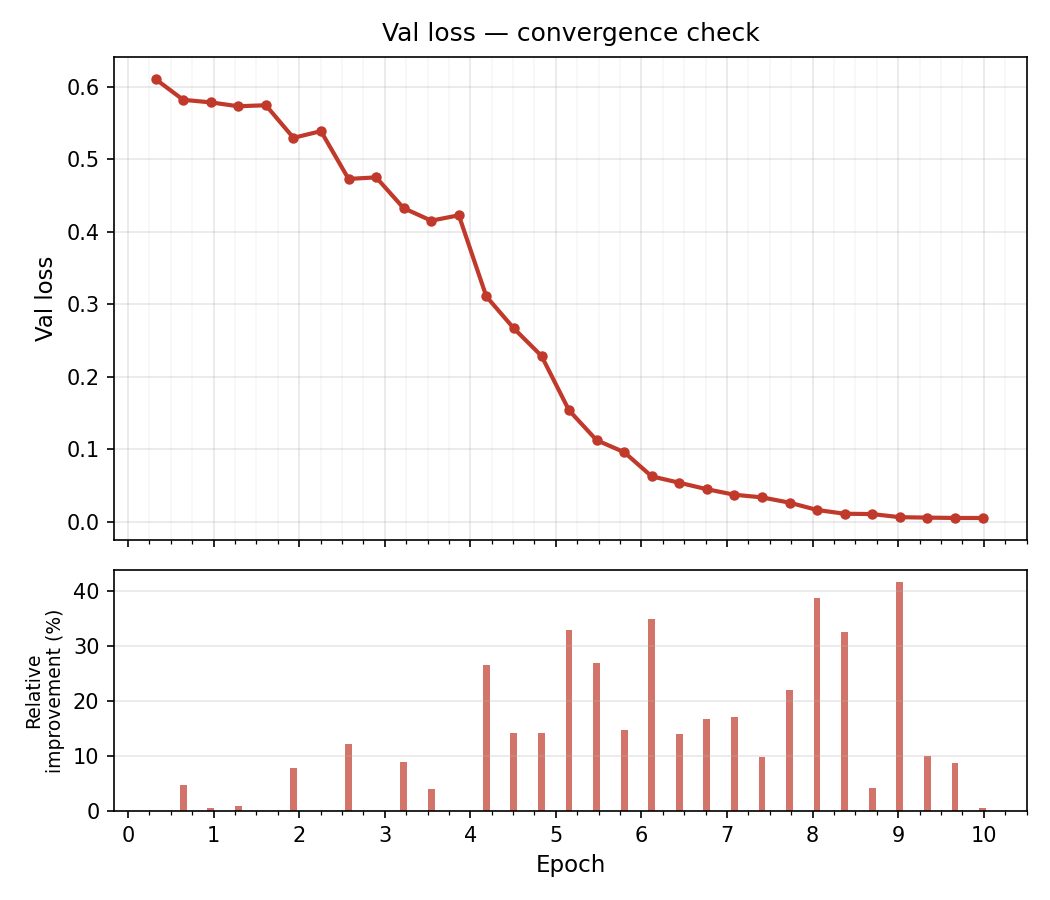}
\caption{Convergence analysis. Validation loss over 10 epochs at
  $\eta{=}2\times10^{-4}$ (top) and per-checkpoint relative improvement
  \% (bottom).}
\label{fig:convergence_targeted}
\end{figure}

Before committing to a full hyperparameter search we trained a single targeted-LoRA model for 10 epochs at a nominal learning rate of $2\times10^{-4}$ to characterise the shape of the learning curve. Figure~\ref{fig:convergence_targeted} shows that training exhibits a slow
initial phase through epoch~4, during which validation loss decreases gradually from $0.61$ to $0.42$. A steep learning phase begins at epoch~4 and persists through epoch~9, with validation loss falling from $0.42$ to approximately $0.01$ and per-checkpoint improvements of $14$--$41\%$.
From epoch~9 onward improvement drops below $10\%$, indicating plateau. This confirmed that a 10-epoch budget is sufficient and motivated the use of early stopping in the subsequent search.
 
\paragraph{Per-window learning-rate search.}
Because the optimal learning rate depends on the number of trainable parameters, and different LoRA windows expose different parameter counts, we conducted an independent log-uniform random search for each window rather than re-using a single shared rate. Five learning rates were sampled from $\eta \sim \mathrm{LogUniform}[10^{-5},\,4\times10^{-4}]$ with a fixed
seed (42) for reproducibility. All other settings matched Table~\ref{tab:fixed-hparams}. The best-validation-loss trial was selected for each window before examining any test performance.

\paragraph{Training data.}
All experiments use the MedArabBench MCQA train split~\citep{daoud2026medarabench}
as the sole training source, comprising 17,860 training and 1,987
validation examples after a stratified 90/10 split (seed 42).



\begin{table*}[!t]
\centering
\small

\setlength{\tabcolsep}{5pt}
\renewcommand{\arraystretch}{1.10}

\begin{tabular}{>{\raggedright\arraybackslash}m{2.5cm}
                l
                c c c}
\specialrule{1pt}{0pt}{1.5pt}

\multicolumn{1}{c}{\textbf{Category}} & \multicolumn{1}{c}{\textbf{Model}} & \multicolumn{3}{c}{\textbf{MedAraBench-OE}} \\
\cmidrule(lr){3-5}
 & & \textbf{BERTScore-F1} & \textbf{Correct \%} & \textbf{Incorrect \%} \\
\specialrule{1pt}{0pt}{1.5pt}
\multirow{3}{2.5cm}{\parbox[c]{2.5cm}{\centering Closed-Source\\General-Purpose}}
& GPT-5.2 & \underline{63.9} & \underline{62.6} & \underline{37.3} \\
& Gemini-2.5-Flash & \textbf{65.1} & 60.5 & 39.5 \\
& Claude-Opus-4.6 & 61.0 & \textbf{67.6} & \textbf{32.3} \\
\specialrule{0.6pt}{0pt}{0pt}
\multirow{6}{2.5cm}{\parbox[c]{2.5cm}{\centering Open-Source\\General-Purpose}}
& Mistral-7B-Instruct-v0.3 & 51.5 & 2.3 & 97.6 \\
& Llama-3.1-8B-Instruct & 54.3 & 6.6 & 93.4 \\
& Mistral-Small-3.2-24B & \underline{60.0} & 30.1 & 69.8 \\
& Gemma-3-27B-IT & 59.3 & \underline{32.6} & \underline{67.3} \\
& Llama-3.3-70B & 57.8 & 19.5 & 80.4 \\
& DeepSeek-V3.2 & \textbf{62.0} & \textbf{49.0} & \textbf{50.9} \\
\specialrule{0.6pt}{0pt}{0pt}
\multirow{6}{2.5cm}{\parbox[c]{2.5cm}{\centering Arabic/Multilingual\\General-Purpose}}
& Jais-2-8B-Chat  & 54.2 & \underline{17.7} & \underline{82.3} \\
& ALLaM-7B-Instruct-Preview  & 56.9 & \textbf{19.6} & \textbf{80.3} \\
& Aya-Expanse-8B & 53.9 & 15.7 & 84.2 \\
& SILMA-9B-Instruct-v1.0  & \underline{57.4} & 15.0 & 85.0 \\
& Falcon-H1-7B-Instruct  & \textbf{57.6} & 15.5 & 84.5 \\
& Fanar-1-9B-Instruct & 48.4 & 7.4 & 92.5 \\
\specialrule{0.6pt}{0pt}{0pt}
\multirow{3}{2.5cm}{\parbox[c]{2.5cm}{\centering Medical-Domain}}
& MedGemma-27B-Text-IT & \textbf{59.8} & \textbf{32.9} & \textbf{67.0} \\
& Meditron-3-70B & \underline{54.7} & \underline{30.7} & \underline{69.3} \\
& Llama-3-Med42-70B & 53.4 & 24.6 & 75.4 \\
\specialrule{0.6pt}{0pt}{0pt}
\multirow{6}{2.5cm}{\parbox[c]{2.5cm}{\centering Adaptation\\Methods}}
& Mistral + Few-Shot (k=5) & \textbf{60.2} & \underline{28.5} & \underline{71.4} \\
& Mistral + AUTOCAP & \underline{59.8} & 27.5 & 72.5 \\
& BiMedix & 47.6 & 18.7 & 81.3 \\
& Mistral + Full LoRA & 53.8 & 25.5 & 75.4 \\
\rowcolor{gray!25}& Mistral + TLoRA (Ours) & 57.9 & \textbf{29.7} & \textbf{70.3} \\
& Mistral + TLoRA (Optimal) & 52.6 & 38.1 & 61.9 \\
\specialrule{1pt}{0pt}{0pt}
\end{tabular}%
\caption{Short Answer Generation results on MedAraBench-OE. 
Models are evaluated using BERTScore-F1 and LLM-as-a-judge correctness labels, reported as the percentage of
correct (Correct \%) and incorrect (Incorrect \%) responses. Models are grouped by family. Bold indicates the best result within each group per column, while underlining points to second-best result. Mistral-Small-3.2-24B-Instruct-2506 is the Mistral backbone used in the adaptation methods.}
\label{tab:task2-results}
\end{table*}

\section{Complete Results}
\label{app:complete-results}

This appendix presents the complete per-model results for the Short Answer Generation and Multi-Turn Clinical Dialogue tasks. All results are evaluated using BERTScore-F1 with AraBERTv2 as the underlying contextual encoder, and an LLM-as-a-judge setup reporting the percentage of correct (Correct~\%) and incorrect (Incorrect~\%) responses.

\subsection{Short Answer Generation Complete Results}
\label{app:complete-results-task2}
Complete results for the Short Answer Generation task are presented in Table~\ref{tab:task2-results}, covering models from all five categories evaluated on MedAraBench-OE with gold answers provided as references to the judge.

\subsection{Multi-Turn Clinical Dialogue Complete Results}
\label{app:complete-results-task3}
Complete results for the Multi-Turn Clinical Dialogue task are presented in Tables~\ref{tab:task3-msa-results}--\ref{tab:task3-egyptian-results}, with one table per AraClinicDialog variant: MSA, Emirati, Moroccan Darija, Jordanian, and Egyptian.







\begin{table*}[!t]
\centering
\small
\setlength{\tabcolsep}{5pt}
\renewcommand{\arraystretch}{1.10}

\begin{tabular}{>{\raggedright\arraybackslash}m{2.5cm}
                l
                c
                c c}
\specialrule{1pt}{0pt}{1.5pt}


\multicolumn{1}{c}{\textbf{Category}} & 
\multicolumn{1}{c}{\textbf{Model}} & 
\multicolumn{3}{c}{\textbf{AraClinicDialog (MSA)}} \\
\cmidrule(lr){3-5}


& & 
\textbf{BERTScore-F1} &
\textbf{Correct \%} & 
\textbf{Incorrect \%} \\
\specialrule{1pt}{0pt}{1.5pt}

\multirow{3}{2.5cm}{\parbox[c]{2.5cm}{\centering Closed-Source\\General-Purpose}}
& GPT-5.2 & \underline{56.3} & \textbf{85.0} & \textbf{15.0} \\
& Gemini-2.5-Flash & \textbf{57.9} & 50.0 & 50.0 \\
& Claude-Opus-4.6 & \textbf{57.9} & \underline{74.0} & \underline{26.0} \\
\specialrule{0.6pt}{0pt}{0pt}
\multirow{6}{2.5cm}{\parbox[c]{2.5cm}{\centering Open-Source\\General-Purpose}}
& Mistral-7B-Instruct-v0.3 & 56.9 & 35.0 & 65.0 \\
& Llama-3.1-8B-Instruct & 57.7 & 20.0 & 80.0 \\
& Mistral-Small-3.2-24B & \textbf{58.4} & 44.0 & 56.0 \\
& Gemma-3-27B-it & 57.6 & \underline{49.0} & \underline{51.0} \\
& Llama-3.3-70B & 57.8 & 34.0 & 66.0 \\
& DeepSeek-V3.2 & \underline{58.0} & \textbf{71.0} & \textbf{29.0} \\
\specialrule{0.6pt}{0pt}{0pt}
\multirow{6}{2.5cm}{\parbox[c]{2.5cm}{\centering Arabic/Multilingual\\General-Purpose}}
& Jais-2-8B-Chat & 52.3 & 26.0 & 74.0 \\
& ALLaM-7B-Instruct-preview & \underline{57.7} & \textbf{43.0} & \textbf{57.0} \\
& Aya-Expanse-8B & \textbf{57.8} & \underline{40.0} & \underline{60.0} \\
& SILMA-9B-Instruct-v1.0 & 53.7 & 18.0 & 82.0 \\
& Falcon-H1-7B-Instruct & 56.6 & 39.0 & 61.0 \\
& Fanar-1-9B-Instruct & 43.8 & 11.0 & 89.0 \\
\specialrule{0.6pt}{0pt}{0pt}
\multirow{3}{2.5cm}{\parbox[c]{2.5cm}{\centering Medical-Domain}}
& MedGemma-27B-Text-it & \textbf{58.0} & \textbf{64.0} & \textbf{36.0} \\
& Meditron-3-70B & 57.3 & 28.0 & 72.0 \\
& Llama-3-Med42-70B & \underline{57.5} & \underline{55.0} & \underline{45.0} \\
\specialrule{0.6pt}{0pt}{0pt}
\multirow{5}{2.5cm}{\parbox[c]{2.5cm}{\centering Adaptation\\Methods}}
& Mistral + Few-Shot (k=5) & \textbf{58.2} & 55.0 & 45.0 \\
& Mistral + AUTOCAP & \underline{56.9} & 29.0 & 71.0 \\
& BiMedix & 56.7 & 32.0 & 68.0 \\
& Mistral + Full LoRA & 56.0 & \textbf{68.0} & \textbf{32.0} \\
\rowcolor{gray!25}& Mistral + TLoRA (Ours) & 52.5 & \underline{65.0} & \underline{35.0} \\

\specialrule{1pt}{0pt}{0pt}
\end{tabular}%
\caption{Multi-Turn Clinical Dialogue results on AraClinicDialog (MSA). Models are evaluated using BERTScore-F1 and LLM-as-a-judge scores for correct (\%) and incorrect (\%) outputs. Models are grouped by family. Bold indicates the best result within each group per column, while underline points to second-best result. Mistral-Small-3.2-24B-Instruct-2506 is the Mistral backbone used in the adaptation methods.}
\label{tab:task3-msa-results}
\end{table*}

\begin{table*}[!t]
\centering
\small
\setlength{\tabcolsep}{5pt}
\renewcommand{\arraystretch}{1.1}

\begin{tabular}{>{\raggedright\arraybackslash}m{2.5cm}
                l
                c
                c c}
\specialrule{1pt}{0pt}{1.5pt}

\multicolumn{1}{c}{\textbf{Category}} & 
\multicolumn{1}{c}{\textbf{Model}} & 
\multicolumn{3}{c}{\textbf{AraClinicDialog (Emirati)}} \\
\cmidrule(lr){3-5}


& & 
\textbf{BERTScore-F1} &
\textbf{Correct \%} & 
\textbf{Incorrect \%} \\
\specialrule{1pt}{0pt}{1.5pt}




\multirow{3}{2.5cm}{\parbox[c]{2.5cm}{\centering Closed-Source\\General-Purpose}}
& GPT-5.2 & 56.5 & \textbf{84.0} & \textbf{16.0} \\
& Gemini-2.5-Flash & \textbf{58.4} & 45.0 & 55.0 \\
& Claude-Opus-4.6 & \underline{57.8} & \underline{77.0} & \underline{23.0} \\
\specialrule{0.6pt}{0pt}{0pt}
\multirow{6}{2.5cm}{\parbox[c]{2.5cm}{\centering Open-Source\\General-Purpose}}
& Mistral-7B-Instruct-v0.3 & 55.2 & 33.0 & 67.0 \\
& Llama-3.1-8B-Instruct & 55.4 & 14.0 & 86.0 \\
& Mistral-Small-3.2-24B & \underline{57.5} & \underline{40.0} & \underline{60.0} \\
& Gemma-3-27B-it & 56.9 & 39.0 & 61.0 \\
& Llama-3.3-70B & 56.7 & 23.0 & 77.0 \\
& DeepSeek-V3.2 & \textbf{57.8} & \textbf{64.0} & \textbf{36.0} \\
\specialrule{0.6pt}{0pt}{0pt}
\multirow{6}{2.5cm}{\parbox[c]{2.5cm}{\centering Arabic/Multilingual\\General-Purpose}}
& Jais-2-8B-Chat & 53.2 & 27.0 & 73.0 \\
& ALLaM-7B-Instruct-preview & 53.9 & 28.0 & 72.0 \\
& Aya-Expanse-8B & \textbf{56.2} & \underline{32.0} & \underline{68.0} \\
& SILMA-9B-Instruct-v1.0 & 50.6 & 15.0 & 85.0 \\
& Falcon-H1-7B-Instruct & \underline{54.9} & \textbf{37.0} & \textbf{63.0} \\
& Fanar-1-9B-Instruct & 45.6 & 9.0 & 91.0 \\
\specialrule{0.6pt}{0pt}{0pt}
\multirow{3}{2.5cm}{\parbox[c]{2.5cm}{\centering Medical-Domain}}
& MedGemma-27B-Text-it & \underline{58.3} & \underline{43.0} & \underline{57.0} \\
& Meditron-3-70B & 56.5 & 38.0 & 62.0 \\
& Llama-3-Med42-70B & \textbf{60.8} & \textbf{58.0} & \textbf{42.0} \\
\specialrule{0.6pt}{0pt}{0pt}
\multirow{5}{2.5cm}{\parbox[c]{2.5cm}{\centering Adaptation\\Methods}}
& Mistral + Few-Shot (k=5) & \textbf{58.3} & \underline{47.0} & \underline{53.0} \\
& Mistral + AUTOCAP & 54.9 & 25.0 & 75.0 \\
& BiMedix & \underline{55.1} & 29.0 & 71.0 \\
& Mistral + Full LoRA & 54.4 & \textbf{49.0} & \textbf{51.0} \\
\rowcolor{gray!25}& Mistral + TLoRA (Ours) & 49.4 & 42.0 & 58.0 \\

\specialrule{1pt}{0pt}{0pt}
\end{tabular}%
\caption{Multi-Turn Clinical Dialogue results on AraClinicDialog (Emirati). Models are evaluated using BERTScore-F1 and LLM-as-a-judge scores for correct (\%) and incorrect (\%) outputs. Models are grouped by family. Bold indicates the best result within each group per column, while underline indicates second-best result. Mistral-Small-3.2-24B-Instruct-2506 is the Mistral backbone used in the adaptation methods.}
\label{tab:task3-emirati-results}
\end{table*}

\begin{table*}[!t]
\centering
\small
\setlength{\tabcolsep}{5pt}
\renewcommand{\arraystretch}{1.1}

\begin{tabular}{>{\raggedright\arraybackslash}m{2.5cm}
                l
                c
                c c}
\specialrule{1pt}{0pt}{1.5pt}

\multicolumn{1}{c}{\textbf{Category}} & 
\multicolumn{1}{c}{\textbf{Model}} & 
\multicolumn{3}{c}{\textbf{AraClinicDialog (Moroccan Darija)}} \\
\cmidrule(lr){3-5}


& & 
\textbf{BERTScore-F1} &
\textbf{Correct \%} & 
\textbf{Incorrect \%} \\
\specialrule{1pt}{0pt}{1.5pt}




\multirow{3}{2.5cm}{\parbox[c]{2.5cm}{\centering Closed-Source\\General-Purpose}}
& GPT-5.2 & 54.6 & \textbf{76.0} & \textbf{24.0} \\
& Gemini-2.5-Flash & \underline{56.2} & 52.0 & 48.0 \\
& Claude-Opus-4.6 & \textbf{56.5} & \underline{70.0} & \underline{30.0} \\
\specialrule{0.6pt}{0pt}{0pt}
\multirow{6}{2.5cm}{\parbox[c]{2.5cm}{\centering Open-Source\\General-Purpose}}
& Mistral-7B-Instruct-v0.3 & 52.9 & 22.0 & 78.0 \\
& Llama-3.1-8B-Instruct & 53.8 & 14.0 & 86.0 \\
& Mistral-Small-3.2-24B & 55.0 & 32.0 & 68.0 \\
& Gemma-3-27B-it & 55.1 & \underline{33.0} & \underline{67.0} \\
& Llama-3.3-70B & \underline{55.4} & 18.0 & 82.0 \\
& DeepSeek-V3.2 & \textbf{56.1} & \textbf{64.0} & \textbf{36.0} \\
\specialrule{0.6pt}{0pt}{0pt}
\multirow{6}{2.5cm}{\parbox[c]{2.5cm}{\centering Arabic/Multilingual\\General-Purpose}}
& Jais-2-8B-Chat & 50.6 & 16.0 & 84.0 \\
& ALLaM-7B-Instruct-preview & \textbf{54.8} & \textbf{36.0} & \textbf{64.0} \\
& Aya-Expanse-8B & \underline{53.5} & \underline{29.0} & \underline{71.0} \\
& SILMA-9B-Instruct-v1.0 & 51.6 & 19.0 & 81.0 \\
& Falcon-H1-7B-Instruct & 53.4 & 21.0 & 79.0 \\
& Fanar-1-9B-Instruct & 45.2 & 11.0 & 89.0 \\
\specialrule{0.6pt}{0pt}{0pt}
\multirow{3}{2.5cm}{\parbox[c]{2.5cm}{\centering Medical-Domain}}
& MedGemma-27B-Text-it & \textbf{55.5} & \underline{40.0} & \underline{60.0} \\
& Meditron-3-70B & \underline{53.9} & 24.0 & 76.0 \\
& Llama-3-Med42-70B & 52.9 & \textbf{44.0} & \textbf{56.0} \\
\specialrule{0.6pt}{0pt}{0pt}
\multirow{5}{2.5cm}{\parbox[c]{2.5cm}{\centering Adaptation\\Methods}}
& Mistral + Few-Shot (k=5) & \textbf{55.7} & 39.0 & 61.0 \\
& Mistral + AUTOCAP & \underline{54.2} & 27.0 & 73.0 \\
& BiMedix & 52.0 & 21.0 & 79.0 \\
& Mistral + Full LoRA & 47.8 & \underline{59.0} & \underline{41.0} \\
\rowcolor{gray!25}& Mistral + TLoRA (Ours) & 48.4 & \textbf{70.0} & \textbf{30.0} \\

\specialrule{1pt}{0pt}{0pt}
\end{tabular}%
\caption{Multi-Turn Clinical Dialogue results on AraClinicDialog (Moroccan Darija). Models are evaluated using BERTScore-F1 and LLM-as-a-judge scores for correct (\%) and incorrect (\%) outputs. Models are grouped by family. Bold indicates the best result within each group per column, while underline points to the second-best result. Mistral-Small-3.2-24B-Instruct-2506 is the Mistral backbone used in the adaptation methods.}
\label{tab:task3-moroccan-darija-results}
\end{table*}

\begin{table*}[!t]
\centering
\small
\setlength{\tabcolsep}{5pt}
\renewcommand{\arraystretch}{1.1}

\begin{tabular}{>{\raggedright\arraybackslash}m{2.5cm}
                l
                c
                c c}
\specialrule{1pt}{0pt}{1.5pt}

\multicolumn{1}{c}{\textbf{Category}} & 
\multicolumn{1}{c}{\textbf{Model}} & 
\multicolumn{3}{c}{\textbf{AraClinicDialog (Jordanian)}} \\
\cmidrule(lr){3-5}


& & 
\textbf{BERTScore-F1} &
\textbf{Correct \%} & 
\textbf{Incorrect \%} \\
\specialrule{1pt}{0pt}{1.5pt}




\multirow{3}{2.5cm}{\parbox[c]{2.5cm}{\centering Closed-Source\\General-Purpose}}
& GPT-5.2 & 56.1 & \textbf{78.0} & \textbf{22.0} \\
& Gemini-2.5-Flash & \textbf{57.7} & 50.0 & 50.0 \\
& Claude-Opus-4.6 & \underline{57.6} & \underline{77.0} & \underline{23.0} \\

\specialrule{0.6pt}{0pt}{0pt}

\multirow{6}{2.5cm}{\parbox[c]{2.5cm}{\centering Open-Source\\General-Purpose}}
& Mistral-7B-Instruct-v0.3 & 54.9 & 32.0 & 68.0 \\
& Llama-3.1-8B-Instruct & 55.6 & 20.0 & 80.0 \\
& Mistral-Small-3.2-24B & 56.0 & 30.0 & 70.0 \\
& Gemma-3-27B-it & \underline{57.1} & \underline{44.0} & \underline{56.0} \\
& Llama-3.3-70B & 56.1 & 26.0 & 74.0 \\
& DeepSeek-V3.2 & \textbf{57.3} & \textbf{67.0} & \textbf{33.0} \\

\specialrule{0.6pt}{0pt}{0pt}

\multirow{6}{2.5cm}{\parbox[c]{2.5cm}{\centering Arabic/Multilingual\\General-Purpose}}
& Jais-2-8B-Chat & 52.6 & 30.0 & 70.0 \\
& ALLaM-7B-Instruct-preview & \underline{55.5} & \textbf{43.0} & \textbf{57.0} \\
& Aya-Expanse-8B & \textbf{56.2} & \underline{34.0} & \underline{66.0} \\
& SILMA-9B-Instruct-v1.0 & 51.1 & 12.0 & 88.0 \\
& Falcon-H1-7B-Instruct & 54.8 & 28.0 & 72.0 \\
& Fanar-1-9B-Instruct & 46.0 & 12.0 & 88.0 \\

\specialrule{0.6pt}{0pt}{0pt}

\multirow{3}{2.5cm}{\parbox[c]{2.5cm}{\centering Medical-Domain}}
& MedGemma-27B-Text-it & \textbf{57.8} & \underline{48.0} & \underline{52.0} \\
& Meditron-3-70B & \underline{56.1} & 33.0 & 67.0 \\
& Llama-3-Med42-70B & 55.2 & \textbf{53.0} & \textbf{47.0} \\

\specialrule{0.6pt}{0pt}{0pt}

\multirow{5}{2.5cm}{\parbox[c]{2.5cm}{\centering Adaptation\\Methods}}
& Mistral + Few-Shot (k=5) & \textbf{56.9} & 40.0 & 60.0 \\
& Mistral + AUTOCAP & \underline{55.4} & 20.0 & 80.0 \\
& BiMedix & 54.9 & 28.0 & 72.0 \\
& Mistral + Full LoRA & 54.2 & \textbf{54.0} & \textbf{46.0} \\
\rowcolor{gray!25}& Mistral + TLoRA (Ours) & 54.5 & \underline{46.0} & \underline{54.0} \\

\specialrule{1pt}{0pt}{0pt}
\end{tabular}%
\caption{Multi-Turn Clinical Dialogue results on AraClinicDialog (Jordanian). Models are evaluated using BERTScore-F1 and LLM-as-a-judge scores for correct (\%) and incorrect (\%) outputs. Bold indicates the best result within each group per column, while underline represents the second-best result. Mistral-Small-3.2-24B-Instruct-2506 is the Mistral backbone used in the adaptation methods.}
\label{tab:task3-jordanian-results}
\end{table*}

\begin{table*}[!t]
\centering
\small
\setlength{\tabcolsep}{5pt}
\renewcommand{\arraystretch}{1.1}
\begin{tabular}{>{\raggedright\arraybackslash}m{2.5cm}
                l
                c
                c c}
\specialrule{1pt}{0pt}{1.5pt}

\multicolumn{1}{c}{\textbf{Category}} & 
\multicolumn{1}{c}{\textbf{Model}} & 
\multicolumn{3}{c}{\textbf{AraClinicDialog (Egyptian)}} \\
\cmidrule(lr){3-5}


& & 
\textbf{BERTScore-F1} &
\textbf{Correct \%} & 
\textbf{Incorrect \%} \\
\specialrule{1pt}{0pt}{1.5pt}


\multirow{3}{2.5cm}{\parbox[c]{2.5cm}{\centering Closed-Source\\General-Purpose}}
& GPT-5.2 & 56.7 & \textbf{80.0} & \textbf{20.0} \\
& Gemini-2.5-Flash & \textbf{58.5} & 50.0 & 50.0 \\
& Claude-Opus-4.6 & \underline{58.4} & \underline{74.0} & \underline{26.0} \\
\specialrule{0.6pt}{0pt}{0pt}
\multirow{6}{2.5cm}{\parbox[c]{2.5cm}{\centering Open-Source\\General-Purpose}}
& Mistral-7B-Instruct-v0.3 & 55.3 & 28.0 & 72.0 \\
& Llama-3.1-8B-Instruct & 56.3 & 14.0 & 86.0 \\
& Mistral-Small-3.2-24B & 56.8 & 29.0 & 71.0 \\
& Gemma-3-27B-it & \underline{58.1} & \underline{46.0} & \underline{54.0} \\
& Llama-3.3-70B & 57.6 & 24.0 & 76.0 \\
& DeepSeek-V3.2 & \textbf{58.6} & \textbf{65.0} & \textbf{35.0} \\
\specialrule{0.6pt}{0pt}{0pt}
\multirow{6}{2.5cm}{\parbox[c]{2.5cm}{\centering Arabic/Multilingual\\General-Purpose}}
& Jais-2-8B-Chat & 53.7 & 25.0 & 75.0 \\
& ALLaM-7B-Instruct-preview & \underline{56.9} & \textbf{38.0} & \textbf{62.0} \\
& Aya-Expanse-8B & \textbf{57.3} & 32.0 & 68.0 \\
& SILMA-9B-Instruct-v1.0 & 51.7 & 14.0 & 86.0 \\
& Falcon-H1-7B-Instruct & 55.8 & \underline{33.0} & \underline{67.0} \\
& Fanar-1-9B-Instruct & 47.2 & 12.0 & 88.0 \\
\specialrule{0.6pt}{0pt}{0pt}
\multirow{3}{2.5cm}{\parbox[c]{2.5cm}{\centering Medical-Domain}}
& MedGemma-27B-Text-it & \textbf{58.4} & \underline{44.0} & \underline{56.0} \\
& Meditron-3-70B & \underline{57.6} & 28.0 & 72.0 \\
& Llama-3-Med42-70B & 55.8 & \textbf{51.0} & \textbf{49.0} \\
\specialrule{0.6pt}{0pt}{0pt}
\multirow{5}{2.5cm}{\parbox[c]{2.5cm}{\centering Adaptation\\Methods}}
& Mistral + Few-Shot (k=5) & \textbf{57.5} & 38.0 & 62.0 \\
& Mistral + AUTOCAP & \underline{56.2} & 21.0 & 79.0 \\
& BiMedix & 55.5 & 29.0 & 71.0 \\
& Mistral + Full LoRA & 54.9 & \underline{39.0} & \underline{61.0} \\
\rowcolor{gray!25}& Mistral + TLoRA (Ours) & 50.4 & \textbf{48.0} & \textbf{52.0} \\
\specialrule{1pt}{0pt}{0pt}
\end{tabular}%
\caption{Multi-Turn Clinical Dialogue results on AraClinicDialog (Egyptian). Models are evaluated using BERTScore-F1 and LLM-as-a-judge scores for correct (\%) and incorrect (\%). Bold indicates the best result within each group per column, while underline points to the second-best result. Mistral-Small-3.2-24B-Instruct-2506 is the Mistral backbone model used in the adaptation methods.}
\label{tab:task3-egyptian-results}
\end{table*}

\section{Additional Results} 
\label{app:add-results}

\subsection{KL Probe Layer Sensitivity}
\label{sec:kl-sensitivity}

The KL alignment objective requires choosing a \emph{probe layer} $\ell_p$ at which the logit-lens KL divergence between the Arabic and English representations is measured.
Following the zone analysis in \S\ref{sec:mechanistic}, we parameterize this choice via
$\tau = \mu + c\sigma$ (with $\mu$, $\sigma$ the mean and standard deviation of per-layer KL across the model), yielding:\\
$c{=}0 \Rightarrow \ell_p{=}29$ (ramp zone),\\
$c{=}1 \Rightarrow \ell_p{=}34$ (active-zone onset, default),\\
$c{=}2 \Rightarrow \ell_p{=}40$ (deep active zone).\\
 
Table~\ref{tab:c-ablation} reports results for the winning window L1--34 at the tuned learning rate $\eta^*{=}2.76\times10^{-5}$, varying only $c$. The default $c{=}1$ ($\ell_p{=}34$) achieves the best or joint-best score on all MCQA benchmarks, validating the zone-threshold heuristic
for probe placement.

\subsection{Effect of LoRA Layer vs.\ KL Alignment}
\label{app:layer-vs-kl}

To disentangle whether gains over the full-LoRA baseline arise from \emph{where} adapters are placed or from the \emph{KL alignment loss}, we compare CE-only and CE+KL variants within each window. 

Table~\ref{tab:ce-only-windows} shows results for CE-only training across all LoRA windows with per-window tuned learning rates. L1--34 leads on MCQA overall, with the closest competitor L1--24 winning only on AraSTEM (64.7 vs.\ 63.1), likely reflecting its stronger coverage of lower layers where morphological features are processed. Windows restricted to upper layers consistently underperform or degrade below zero-shot, and full-model LoRA collapses entirely at its tuned rate,
producing near-random MCQA scores alongside anomalously high generation BERTScores that LLM-judge evaluation confirms as degenerate repetitive outputs. These results establish that the performance advantage of L1--34 holds even without the KL alignment term, pointing to layer placement as the
primary driver.

\subsection{Fixed Learning Rate Comparison}
\label{app:fixed-lr}

A potential confound in comparing LoRA windows is that each uses a different tuned learning rate.
To isolate layer placement from optimisation, we retrain every window with CE+KL loss at a single fixed rate ($\eta{=}1.10\times10^{-5}$, the tuned optimum for L1--34). Table~\ref{tab:same-lr} shows that L1--34 retains its lead on MedAraBench (62.0) and MedarabiQ (60.0), confirming that the window-level ordering is not an artefact of a more favourable learning rate. Taken together with the CE-only results, both loss function and learning rate can be ruled out as confounds, leaving layer selection as the explanation for the observed gains.

\begin{table*}[!htbp]
\centering
\setlength{\tabcolsep}{5pt}
\small
\renewcommand{\arraystretch}{1.2}
\resizebox{\textwidth}{!}{%
\begin{tabular}{lccccccccc}
\toprule
\multirow{2}{*}{\textbf{Probe}} & \multirow{2}{*}{$c$} &
  \multicolumn{4}{c}{\textbf{MCQA}} &
  \multicolumn{2}{c}{\textbf{Generation}} &
  \multicolumn{2}{c}{\textbf{Dialogue (MSA)}} \\
\cmidrule(lr){3-6}\cmidrule(lr){7-8}\cmidrule(lr){9-10}
 & & MedAraBench & MedarabiQ & MMLU-Bio & AraSTEM
   & BERTScore & LLM judge
   & BERTScore & LLM judge \\
\midrule
L29 & 0 & 60.1 & 54 & 60.5 & 60.1 & 54.3 & 21.9 & 57.0 & \textbf{69} \\
\textbf{L34} & \textbf{1} & \textbf{62.1} & \textbf{60} & \textbf{61.6} & \textbf{65.1} & \textbf{57.9} & 29.7 & 52.5 & 64 \\
L40 & 2 & 59.9 & 53 & 60.8 & 62.5 & 55.4 & \textbf{31.1} & 55.3 & \textbf{69} \\
\bottomrule
\end{tabular}
}
\caption{Probe layer sensitivity ($c$-ablation) for the winning window L1--34, $\eta^*{=}2.76\times10^{-5}$. \textbf{Bold} = best per column.}
\label{tab:c-ablation}
\end{table*}

\begin{table*}[!htbp]
\centering
\setlength{\tabcolsep}{5pt}
\renewcommand{\arraystretch}{1.2}
\resizebox{\textwidth}{!}{%
\begin{tabular}{llccccccccc}
\toprule
\multirow{2}{*}{\textbf{Experiment}} & \multirow{2}{*}{\textbf{Layers}} &
  \multirow{2}{*}{\textbf{Tuned LR}} &
  \multicolumn{4}{c}{\textbf{MCQA}} &
  \multicolumn{2}{c}{\textbf{Generation}} &
  \multicolumn{2}{c}{\textbf{Dialogue (MSA)}} \\
\cmidrule(lr){4-7}\cmidrule(lr){8-9}\cmidrule(lr){10-11}
 & & & MedAraBench & MedarabiQ & MMLU-Bio & AraSTEM
   & BERTScore & LLM judge & BERTScore & LLM judge \\
\midrule
Zero-shot & --- & --- & 52.6 & 51.5 & 55.2 & 62.6 & 60.0 & 30.1 & 58.4 & 44.0 \\
\midrule
Full LoRA  & L1--40  & $1.51\times10^{-4}$ & 42.6 & 42.0 & 45.5 & 36.3 & 90.1$^\dagger$ & 34.1 & 31.7 & 28.0 \\
Targeted   & L1--24  & $2.76\times10^{-5}$ & 61.2 & 58.0 & 60.1 & \textbf{64.7} & 46.9 & 33.0 & 57.6 & 70.0 \\
Targeted   & \textbf{L1--34}  & $2.28\times10^{-5}$ & \textbf{62.5} & \textbf{58.0} & \textbf{61.4} & 63.1 & 45.7 & \textbf{35.3} & \textbf{58.3} & \textbf{71.0} \\
Targeted   & L24--40 & $2.76\times10^{-5}$ & 52.9 & 40.0 & 55.0 & 58.5 & 69.0 & 37.6 & 46.7 & 33.0 \\
Targeted   & L34--40 & $1.06\times10^{-4}$ & 49.4 & 36.0 & 51.7 & 51.0 & 87.5$^\dagger$ & 30.7 & 22.2 & 2.0 \\
\bottomrule
\end{tabular}%
}
\caption{CE-only training across LoRA windows with per-window tuned learning rates.
  \textbf{Bold} = best per column excluding zero-shot.
  $^\dagger$Anomalously high generation BERTScore for Full LoRA and L34--40
  reflects degenerate repetitive outputs, confirmed by low LLM-judge and dialogue scores.}
\label{tab:ce-only-windows}
\end{table*}

\begin{table*}[!htbp]
\centering
\setlength{\tabcolsep}{5pt}
\renewcommand{\arraystretch}{1.2}
\resizebox{\textwidth}{!}{%
\begin{tabular}{llccccccccc}
\toprule
\multirow{2}{*}{\textbf{Experiment}} & \multirow{2}{*}{\textbf{Layers}} &
  \multirow{2}{*}{\textbf{LR}} &
  \multicolumn{4}{c}{\textbf{MCQA}} &
  \multicolumn{2}{c}{\textbf{Generation}} &
  \multicolumn{2}{c}{\textbf{Dialogue (MSA)}} \\
\cmidrule(lr){4-7}\cmidrule(lr){8-9}\cmidrule(lr){10-11}
 & & & MedAraBench & MedarabiQ & MMLU-Bio & AraSTEM
   & BERTScore & LLM judge & BERTScore & LLM judge \\
\midrule
Zero-shot & --- & --- & 52.6 & 51.5 & 55.2 & 62.6 & 60.0 & 30.1 & 58.4 & 44.0 \\
\midrule
Full LoRA  & L1--40  & $1.10\times10^{-5}$ & 61.9 & 55.0 & 60.3 & 60.7 & \textbf{53.8} & 25.5 & 56.0 & 68.0 \\
Targeted   & L1--24  & $1.10\times10^{-5}$ & 59.0 & 49.0 & \textbf{60.9} & \textbf{61.9} & \textbf{53.8} & 29.0 & 57.0 & 69.0 \\
Targeted   & L24--40 & $1.10\times10^{-5}$ & 51.4 & 34.0 & 52.5 & 56.1 & 52.2 & \textbf{40.2} & 48.4 & \textbf{81.0} \\
Targeted   & \textbf{L1--34}  & $1.10\times10^{-5}$ & \textbf{62.0} & \textbf{60.0} & 59.6 & 61.0 & 53.7 & 24.9 & \textbf{57.0} & 70.0 \\
Targeted   & L34--40 & $1.10\times10^{-5}$ & 42.5 & 37.0 & 51.3 & 53.5 & 50.1 & 29.8 & 48.2 & 73.0 \\
\bottomrule
\end{tabular}%
}
\caption{Fixed learning-rate comparison ($\eta{=}1.10\times10^{-5}$ for all windows),
  CE+KL loss, probe L34. \textbf{Bold} = best per column excluding zero-shot.}
\label{tab:same-lr}
\end{table*}

\section{Computational Resources}
\label{app:compute}

All experiments were conducted on the NYUAD Jubail HPC cluster using an average of two NVIDIA A100 GPUs over approximately three months of extensive experimentation by two researchers. This corresponds to a heuristic computational budget of approximately $90$ days $\times$ $20$ hours/day $\times$ $2$ A100 GPUs $= 3{,}600$ A100 GPU-hours.

Table~\ref{tab:compute-stages} breaks down the cost of the diagnostic pipeline itself, run once per base model. The pipeline completes in under two hours of wall-clock time (3.63 GPU-hours total), making the layer-selection diagnosis practical to apply to a new base model without incurring substantial additional cost beyond the exploratory experimentation reported.

\begin{table*}[!htbp]
\centering
\small
\setlength{\tabcolsep}{5pt}
\renewcommand{\arraystretch}{1.2}
\begin{tabular}{lccc}
\toprule
\textbf{Stage} & \textbf{Wall-clock} & \textbf{Hardware} & \textbf{GPU-hours} \\
\midrule
Baseline EN + AR zero-shot inference & 1h40m     & 2$\times$ A100-SXM4-80GB & 3.33 \\
Tuned lens (train + eval)            & 10m50s    & 1$\times$ A100-SXM4-80GB & 0.18 \\
Causal activation patching           & 2m10s     & 1$\times$ A100-SXM4-80GB & 0.04 \\
KL-divergence profiling              & 4m32s     & 1$\times$ A100-SXM4-80GB & 0.08 \\
\midrule
Total & 1h57m32s & --- & 3.63 \\
\bottomrule
\end{tabular}
\caption{Per-stage wall-clock time, hardware, and GPU-hours for the diagnostic pipeline (run once per base model).}
\label{tab:compute-stages}
\end{table*}

\section{Human Annotators}
Ten student annotators contributed to inter-annotator agreement evaluation and medical data review. Annotators were compensated via honoraria and Amazon vouchers commensurate with the hours contributed. We estimate a total annotation effort of approximately $60$ hours across all annotators.

\section{Generalizability of TLoRA}
\label{app:generalizability}

To assess whether TLoRA generalizes beyond the Mistral backbone used in the main text, we apply the same diagnostic pipeline to a different model family and scale, Llama-3.1-8B-Instruct.

Mechanistic analysis on this model identifies $L_{\text{patch}} = $~L18 and $L_{\text{KL}} = $~L28, which define the five candidate windows evaluated below. All windows are trained with the same learning rate ($1.06\times10^{-4}$), selected as the best value for full LoRA.

Table~\ref{tab:generalizability} reports results for each window on Llama-3.1-8B-Instruct (best result per column in bold, second-best starred). Window w1 (layers L1--L18) is selected as TLoRA based on its MCQA performance.

\begin{table*}[!htbp]
\centering
\setlength{\tabcolsep}{5pt}
\renewcommand{\arraystretch}{1.2}
\resizebox{\textwidth}{!}{%
\begin{tabular}{llccccccccc}
\toprule
\multirow{2}{*}{\textbf{Window}} & \multirow{2}{*}{\textbf{LoRA Layers}} &
  \multirow{2}{*}{\textbf{LR}} &
  \multicolumn{4}{c}{\textbf{MCQA}} &
  \multicolumn{2}{c}{\textbf{Generation}} &
  \multicolumn{2}{c}{\textbf{Dialogue (MSA)}} \\
\cmidrule(lr){4-7}\cmidrule(lr){8-9}\cmidrule(lr){10-11}
 & & & MedAraBench & MedarabiQ & MMLU-Bio & AraSTEM
   & BERTScore & LLM judge & BERTScore & LLM judge \\
\midrule
Zero-shot & --- & --- & 37.7 & 32.6 & 38.6 & 37.8 & 54.3 & 6.6 & 57.7 & 20.0 \\
\midrule
Full LoRA & L1--L32  & $1.06\times10^{-4}$ & \textbf{49.6} & 39.0 & 44.6 & 37.4 & 51.7 & 9.6 & 47.7 & \textbf{26.0} \\
Targeted             & \textbf{L1--L18}  & $1.06\times10^{-4}$ & 47.3 & 39.0 & \textbf{44.7} & \textbf{39.5} & \textbf{52.0} & 13.3 & \textbf{50.0} & 25.0 \\
Targeted              & L18--L32 & $1.06\times10^{-4}$ & 41.1 & 36.0 & 39.3 & 33.1 & 46.3 & 6.1 & 44.2 & 22.0 \\
Targeted              & L1--L28  & $1.06\times10^{-4}$ & 46.6 & \textbf{42.0} & 43.7 & 38.1 & \textbf{52.0} & \textbf{13.5} & 46.9 & 13.0 \\
Targeted              & L28--L32 & $1.06\times10^{-4}$ & 34.6 & 23.0 & 37.5 & 31.6 & 46.4 & 4.7 & 21.2 & 1.0 \\
\bottomrule
\end{tabular}%
}
\caption{Generalizability of TLoRA to Llama-3.1-8B-Instruct, fixed learning rate ($\eta{=}1.06\times10^{-4}$ for all windows). \textbf{Bold} = best per column excluding zero-shot. w1 (L1--L18) is selected as TLoRA.}
\label{tab:generalizability}
\end{table*}

TLoRA improves over both the zero-shot baseline and full LoRA on all four MCQA datasets, and achieves the best result on MMLU-Bio (44.7) and AraSTEM (39.5); full LoRA is slightly ahead on MedAraBench (49.6 vs.\ 47.3). On the out-of-domain generation and dialogue tasks, TLoRA is the second-best configuration on generation BERTScore (52.0) and LLM-judge correctness (13.3), and on dialogue LLM-judge correctness (25.0).

\section{Qualitative Case Studies}
\label{app:qualitative-case-studies}

We conduct a paired qualitative comparison between TLoRA and base Mistral-Small-3.2-24B-Instruct-2506 model on the MCQA task. We categorize predictions into four outcome types: \textit{successfully recovered} (incorrect under the base model, correct under TLoRA), \textit{flipped to incorrect} (correct under the base model, incorrect under TLoRA), \textit{stayed correct}, and \textit{failed to recover} (incorrect under both). Cases are pooled across the four MCQA datasets, and we manually review 30 success cases and 30 failure cases.

TLoRA is most beneficial on single-fact retrieval questions, as illustrated by the recovered example in Table~\ref{tab:qualitative}. It struggles, however, to distinguish between closely related answer options: in the failed-to-recover example, both models converge on the same plausible distractor rather than the correct answer. More broadly, many failures involve replacing one incorrect answer with another, and negative or exclusion wording in the question does not appear to be a primary driver of failure cases. An extended table covering all four outcome categories across the MCQA, generation, and dialogue tasks is provided in Appendix~J.

\begin{table*}[t]
\centering
\footnotesize
\renewcommand{\arraystretch}{1.3}
\setlength{\tabcolsep}{4pt}
\resizebox{\textwidth}{!}{%
\begin{tabular}{>{\RaggedRight\arraybackslash}p{1.6cm}
                >{\RaggedRight\arraybackslash}p{1.6cm}
                >{\RaggedRight\arraybackslash}p{7cm}
                >{\RaggedRight\arraybackslash}p{2.4cm}
                >{\RaggedRight\arraybackslash}p{2.4cm}
                >{\RaggedRight\arraybackslash}p{2.4cm}}
\specialrule{1pt}{0pt}{1.5pt}
\textbf{Case Type} & \textbf{Dataset} & \textbf{Question} & \textbf{Mistral-Small} & \textbf{TLoRA} & \textbf{Ground Truth} \\
\specialrule{1pt}{1.5pt}{1.5pt}
Flipped to correct & MedArabiQ &
\textarabic{التبادل المتبادل بين الكروموسوم 8 و14 يسبب: A. لمفوما مانتيل; B. لمفوما بوركت; C. لمفوما هودجكن; D. نقيوم متعدد; E. ابيضاض لمفاوي حاد} — A reciprocal translocation between chromosomes 8 and 14 causes: A.\ Mantle cell lymphoma; B.\ Burkitt lymphoma; C.\ Hodgkin lymphoma; D.\ Multiple myeloma; E.\ Acute lymphoblastic leukemia
& A. Mantle cell lymphoma & B. Burkitt lymphoma & B. Burkitt lymphoma \\
\addlinespace
Stayed incorrect & AraSTEM &
\textarabic{امرأة عمرها 45 عاماً لديها ألم في الأصابع عند التعرض للبرد وآلام مفاصل وصعوبة في بلع الأطعمة الصلبة، إن الفحص المفضل لإجراء تشخيص نهائي هو: A. العامل الرثوي; B. أضداد مضادة للنوى; C. تخطيط قلب كهربائي; D. نيتروجين يوريا الدم والكرياتينين} — A 45-year-old woman has finger pain when exposed to cold, joint pain, and difficulty swallowing solid food. The preferred test for a final diagnosis is: A.\ Rheumatoid factor; B.\ Antinuclear antibodies; C.\ Electrocardiogram; D.\ Blood urea nitrogen and creatinine
& A. Rheumatoid factor & A. Rheumatoid factor & B. Antinuclear antibodies \\
\addlinespace
Flipped to incorrect & MedAraBench &
\textarabic{من فروع الشريان السباتي الباطن: A. الشريان العيني; B. الشريان المخيخي السفلي الأمامي; C. الشريان المخيخي السفلي الخلفي; D. الشريان الشوكي} — Which of the following is a branch of the internal carotid artery? A.\ Ophthalmic artery; B.\ Anterior inferior cerebellar artery; C.\ Posterior inferior cerebellar artery; D.\ Spinal artery
& A. Ophthalmic artery & D. Spinal artery & A. Ophthalmic artery \\
\specialrule{1pt}{0pt}{0pt}
\end{tabular}%
}
\caption{Representative case studies comparing TLoRA and base Mistral-Small-3.2-24B-Instruct-2506 (Mistral-Small) on MCQA. Arabic question stems are shown with their English translation.}
\label{tab:qualitative}
\end{table*}

  

\end{document}